\documentclass[12pt, a4paper]{article}
\usepackage[margin=0.8in]{geometry}
\usepackage{array, multirow}
\usepackage{makecell}
\usepackage{textcomp}
\usepackage{xspace}
\usepackage{amsmath}
\usepackage{graphicx}

\usepackage[round]{natbib} 

\usepackage{tikz}
\let\svtikzpicture\tikzpicture
\def\tikzpicture{\noindent\svtikzpicture}

\usepackage{pgfplots}
\usetikzlibrary{trees}

\usepackage{rotating}
\usepackage{subfig}

\usepackage{hyperref,xcolor}

\usepackage{lscape}
\usepackage[colorinlistoftodos,textsize=scriptsize]{todonotes}

\providecommand{\keywords}[1]
{
  \small	
  \textbf{\textit{Keywords---}} #1
}

\newcommand{\Cross}{\mathbin{\tikz [x=1.4ex,y=1.4ex,line width=.2ex] \draw (0,0) -- (1,1) (0,1) -- (1,0);}}%
\def\Tick{\tikz\fill[scale=0.4](0,.35) -- (.25,0) -- (1,.7) -- (.25,.15) -- cycle;} %
\def\Ticktext{\tikz\fill[scale=0.3](0,.35) -- (.25,0) -- (1,.7) -- (.25,.15) -- cycle;} %

\title{\LARGE Deep Learning Based HEp-2 Image Classification: A Comprehensive Review}

\author{{\normalsize Saimunur Rahman$^{1,2}$, Lei Wang$^1$\footnote{Co-corresponding authors} , Changming Sun$^{2*}$, Luping Zhou$^3$}}
\date{%
    \small$^1$VILA, School of Computing and Information Technology, University of Wollongong, NSW 2522, Australia\\%
    $^2$CSIRO Data61, PO Box 76, Epping, NSW 1710, Australia\\%
    $^3$School of Electrical and Information Engineering, University of Sydney, NSW 2006, Australia\\[2ex]%
    \texttt{\footnotesize \href{mailto:sr801@uowmail.edu.au}{sr801@uowmail.edu.au}, \href{mailto:leiw@uow.edu.au}{leiw@uow.edu.au}, \href{mailto:changming.sun@csiro.au}{changming.sun@csiro.au}, \href{mailto:luping.zhou@sydney.edu.au}{luping.zhou@sydney.edu.au}}
}

\begin{document}
\maketitle




\begin{abstract}
Classification of HEp-2 cell patterns plays a significant role in the indirect immunofluorescence test for identifying autoimmune diseases in the human body. Many automatic HEp-2 cell classification methods have been proposed in recent years, amongst which deep learning based methods have shown impressive performance. This paper provides a comprehensive review of the existing deep learning based HEp-2 cell image classification methods. These methods perform HEp-2 image classification at two levels, namely, cell-level and specimen-level. Both levels are covered in this review. At each level, the methods are organized with a deep network usage based taxonomy. The core idea, notable achievements, and key strengths and weaknesses of each method are critically analyzed. Furthermore, a concise review of the existing HEp-2 datasets that are commonly used in the literature is given. The paper ends with a discussion on novel opportunities and future research directions in this field. It is hoped that this paper would provide readers with a thorough reference of this novel, challenging, and thriving field.
\end{abstract}




\section{Introduction}
Indirect immunofluorescence (IIF) is widely recognized as the gold standard test for the characterization of autoimmune diseases such as rheumatoid arthritis, pulmonary fibrosis, Sjogren's syndrome, and Addison disease in the human body \citep{fritzler1986immunofluorescent, hiemann2007automatic, manivannan2016automated}. IIF is applied to the blood serum, and auto-antibodies are spotted from the fluorescence patterns present in the humane elliptical 2 (HEp-2) cells \citep{hiemann2007automatic}. HEp-2 cells characterise more than thirty cytoplasmic and nuclear patterns existing in almost one hundred types of auto-antibodies \citep{perner2002mining}. However, only a few types of staining patterns are useful for diagnostic purposes, namely, homogeneous, nucleolar, centromereseen, cytoplasmatic, fine speckled, coarse speckled, cytoplasmatic, nucleolar membrane, and golgi. The classification of these patterns from HEp-2 cells remains a challenging task, primarily due to the subtle category differences and the lack of image acquisition standardization  \citep{liu2014hep}. Traditionally, the classification of the patterns is carried out manually by specialist physicians or pathologists. Specifically, they observe every cell in slide under the microscope, and recognize the patterns based on their experience in the field. Nevertheless, despite the long time requirement for the test, the results are not consistent among laboratories due to inter-observer disagreements \citep{gao2017hep}. Therefore, to mitigate these problems and standardize the traditional IIF test practice, the design of reliable and automated HEp-2 image classification systems has become an active area of research.

\begin{figure*}[t]
    \centering
    \includegraphics[width=1\textwidth]{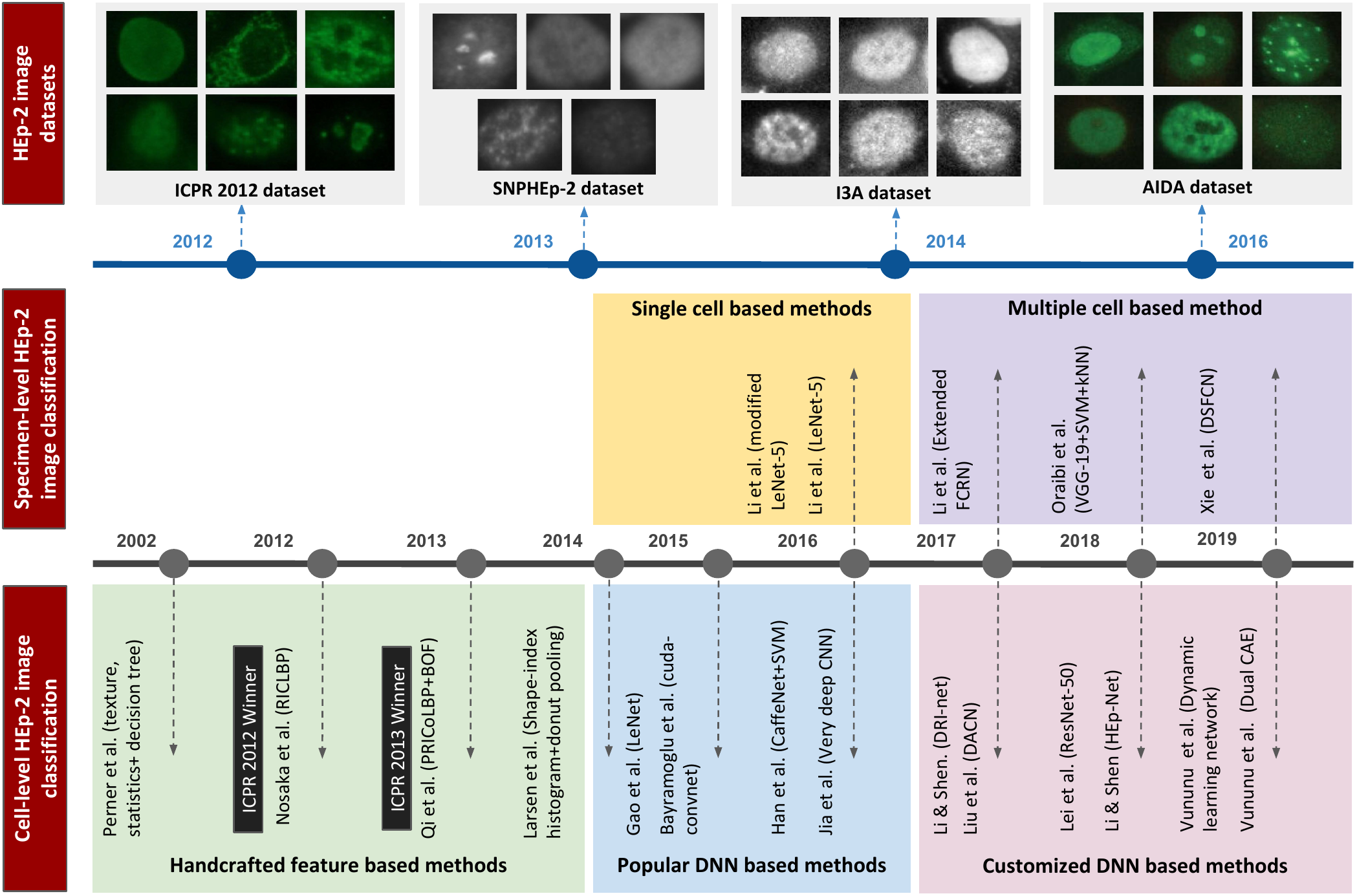}
    \caption{The evolution of DL based HEP2IC methods and benchmark datasets.}
    \label{fig:hepic-evolution}
\end{figure*}

As an established and challenging problem in the field of medical image analysis, HEp-2 image classification (HEP2IC) has been a growing area of research since 2002 \citep{perner2002mining}. Two areas of HEP2IC have received attention of researchers, namely,  individual HEp-2 cell classification and HEp-2 specimen classification. Notable progresses \citep{bayramoglu2015human, ebrahim2018performance, foggia2013benchmarking, gao2017hep, gao2014hep, han2016hep, hobson2015benchmarking, jia2016deep, lei2018deeply, lei2017cross, li2018hep, li2017deep, li2017joint, li2016deep, liu2017hep, lu2017hep,  phan2016transfer, rodrigues2017exploitingnpreprocessing, rodrigues2017hep, shen2018deep} have been made in individual HEp-2 cell classification (also known as cell-level HEP2IC). The HEp-2 specimen classification \citep{li2017hep} (also known as specimen-level HEP2IC) is still a relatively new area of research. Three international IIF image classification competitions were organized in 2012 \citep{foggia2013benchmarking}, 2014 \citep{lovell2014performance}, and 2016 \citep{lovell2016international}, and these competitions play a great role in this progress. As a classical image classification problem, machine learning (ML) techniques have been widely applied to HEP2IC. The advantage of using ML is that it has the capability of learning from data, while non-learning based techniques rely on rules which significantly depend on domain knowledge. However, traditional ML techniques do not directly learn from raw data but rely on predefining some feature representations, which is a critical task and requires complex engineering and a substantial amount of domain experience. A few successful feature representations used by traditional ML techniques for HEP2IC include: rotation invariant local binary pattern \citep{nosaka2014hep}, gradient features with intensity order pooling \citep{shen2014hep},  multiple linear projection descriptors \citep{liu2014hep}, and root-scale invariant feature transform features \& multi-resolution local patterns \citep{manivannan2016automated}.

Deep learning \citep{krizhevsky2012imagenet}, a representation learning approach that automatically learns feature representations from raw data, has received considerable attention from researchers in recent years. It is a very powerful approach that has shown its excellent performance in many areas of medical imaging \citep{litjens2017survey}, including HEP2IC. Among many popular deep neural network (DNN) architectures, the convolutional neural network (CNN) \citep{lecun1998gradient} has been widely used in HEP2IC. A CNN is a supervised learning algorithm that takes labeled images as input and learns robust hierarchical representations which are then used for the classification task. Unsupervised learning algorithms are also employed to learn feature representations. For example, convolutional autoencoder \citep{masci2011stacked} is an unsupervised neural network that has been successfully used for feature learning \citep{liu2017hep} from HEp-2 cell images. The main benefit of using unsupervised learning algorithms is that they do not require image labels.

\begin{figure*}[t]
    \centering
    \includegraphics[width=1\textwidth]{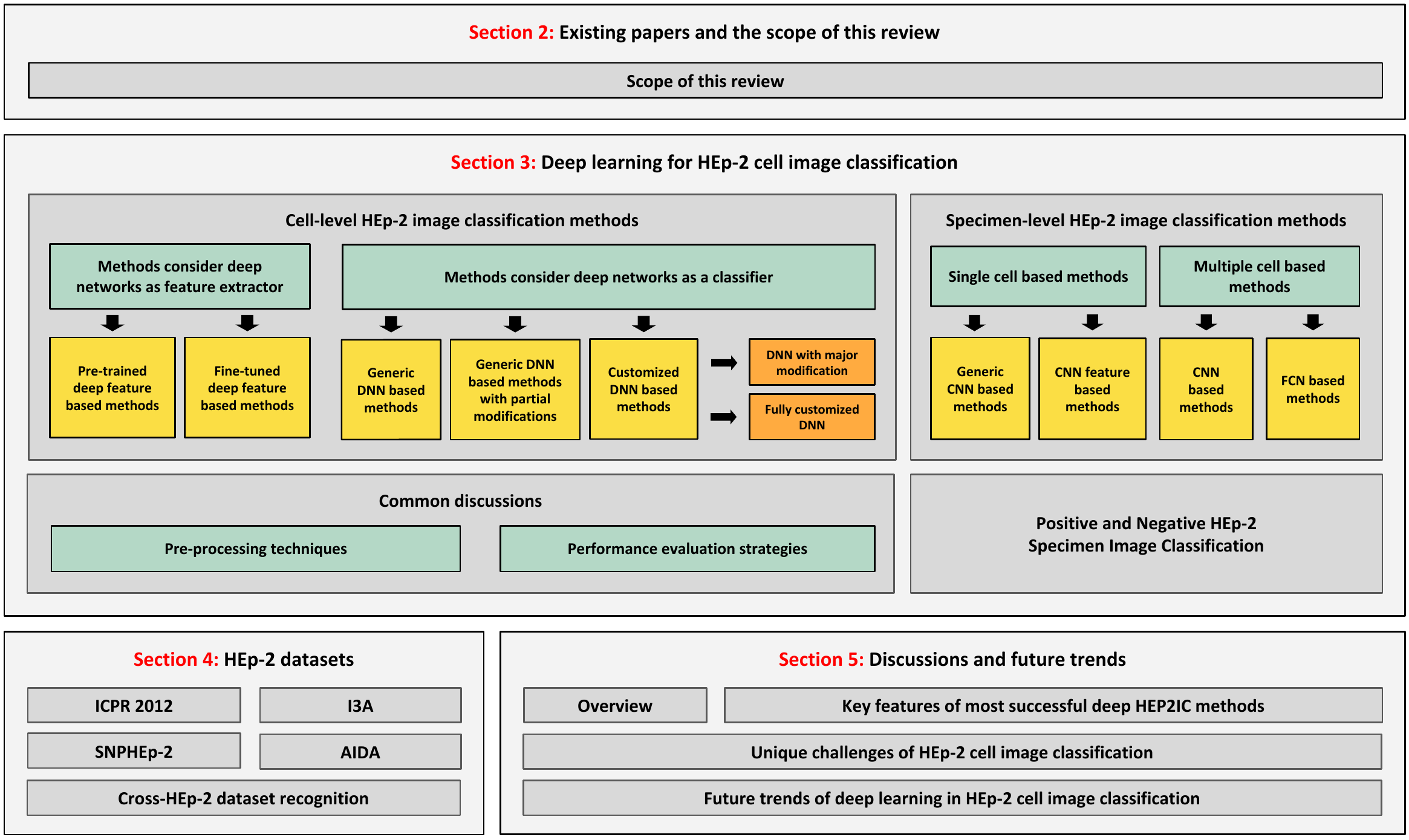}
    \caption{Pictographic overview of the organization of this review. Better viewed in color.}
    \label{fig: taxonomy}
\end{figure*}

The use of deep learning in HEP2IC first began in competitions, then conferences, and very recently in journals. The transition from handcrafted methods to deep learning methods happened during 2013-2014. The size of datasets also emerged from small to medium-scale during that period. Figure \ref{fig:hepic-evolution} shows the evolution of HEP2IC with regard to methods and datasets. In 2016 and 2017, the highest number of deep HEP2IC methods were published. Three dedicated reviews on HEP2IC methods \citep{foggia2014pattern,foggia2013benchmarking,lovell2016international} were published during 2013-2016. These reviews were published as a part of the IIF image classification competitions, and mostly focused on providing the overview of the methods participating in the competitions. There were only a few DL methods involved in these competitions, amongst which \citep{gao2014hep}'s method ranked high. The majority of the participating and winning methods in these competitions were based on traditional handcrafted features (as shown in Figure \ref{fig:hepic-evolution}). It is worth mentioning that there were no new HEP2IC competitions organized after 2016. Two recent reviews \citep{litjens2017survey,xing2017deep} on the broader application of deep learning in microscopic image analysis mention a few HEP2IC methods (as shown in Tables \ref{tab:cell-level methods} and \ref{tab: sl-hep2ic}) as part of their review and discussed the achievements and the pros and cons of these methods. To the best of our knowledge, none of the above reviews has fully covered the publications related to deep learning based HEP2IC methods.

Motivated by the analysis above, this paper aims to provide a comprehensive review of deep learning based HEP2IC methods that have not been discussed by existing reviews. In particular, the state-of-the-art HEP2IC methods published from 2013 to October 2019 (in peer-reviewed conferences and journals, and arXiv) are the focus of this review. The methods are critically reviewed by highlighting the core ideas, pros and cons, and key achievements. For quick reference, the summary of existing methods with key information is also presented in a tabular format. Since datasets are an integral part of HEP2IC methods, a thorough review of the existing HEp-2 datasets is also provided. Based on our experience in deep learning and study on HEP2IC methods, a dedicated section that discusses the key features of current high-performing methods, the open challenges,  and the future trends of HEP2IC research is presented at the end of this review. To summarize, the objective of this review is to:

\begin{enumerate}
\item [(1)] show the recent progresses of HEP2IC based on deep learning,
\item [(2)] discover the key challenges to deep learning based HEP2IC systems,
\item [(3)] manifest the contributions made by researchers to deal with these challenges, and
\item [(4)] highlight the novel opportunities in HEP2IC and the experience gained through recent research.
\end{enumerate}

The rest of the paper is organized as follows. Section 2 provides a comparison of this review with the existing related papers and discusses the scope of this review. Section 3 gives a comprehensive review of existing deep learning based HEP2IC methods. For ease of understanding, the methods are organized in deep network usage based taxonomy. Section 4 gives a concise discussion of the existing public HEp-2 datasets with their evaluation criteria. Finally, Section 5 concludes this review with the discussions on the key features of existing high-performing HEP2IC methods, the unique challenges of HEP2IC, and the expected research trends in the coming years. Figure \ref{fig: taxonomy} gives an overview of the content of the following sections.

\begin{table*}[t]
\scalebox{0.842}{%
\begin{tabular}{llll} 
\hline
 \multirow{2}{*}{\textbf{Reference}} &  \multirow{2}{*}{\textbf{Summary of paper}} & \multicolumn{2}{l}{\textbf{Focused feature domain}}   \\ \cline{3-4}
 &  & \textbf{Handcrafted} & \textbf{Deep learning}  \\ \hline
 \makecell[l]{\cite{foggia2013benchmarking}} & ICPR 2013 IIF image competition methods.  & $\Tick$ &$\Cross$ \\ \hline
\makecell[l]{\cite{foggia2014pattern}} & ICPR 2014 IIF image competition methods.  &  $\Tick$ &$\Tick$ \\ \hline
\makecell[l]{\cite{lovell2016international}} & ICPR 2016 IIF image competition methods.  &  $\Tick$ &$\Cross$ \\ \hline
\makecell[l]{\cite{litjens2017survey}} & DL methods for medial image analysis. &  $\Cross$ &$\Tick$ \\ \hline
\makecell[l]{\cite{xing2017deep}} & DL methods for microscopic image analysis. &  $\Cross$ &$\Tick$ \\ \hline
\makecell[l]{\textbf{This work (2019)}} & DL methods for HEp-2 image classification. &  $\Cross$ &$\Tick$ \\ \hline
\end{tabular}}
\caption{Summary of existing review papers involving HEp-2 cell image classification and deep learning from 2013 to 2019. The symbols $\protect\Ticktext$ and  $\protect\Cross$ indicate that the survey covers and does not cover the papers from a particular domain, respectively.}
\label{tab: existing surveys}
\end{table*}

\section{Existing Papers and the Scope of This Review}
There have been a large number of independent studies conducted for both the deep learning (DL) and the HEP2IC. Studies combining both areas have emerged in the recent years. A few independent reviews covering selected important aspects of either areas are available in the literature. Briefly, the existing DL reviews \citep{guo2016deep,rawat2017deep,zhao2019object,yang2019deep,signoroni2019deep,liu2020deep} have two primary focuses, namely, (1) covering generic DL architectures and discussing their potential applications, and (2) covering DL techniques for specific application domains such as generic image classification. On the other hand, the HEP2IC reviews were published as a part of HEP2IC competitions \citep{foggia2014pattern,foggia2013benchmarking,lovell2016international} mentioned earlier and mostly focus on reviewing the handcrafted methods since the DL methods were limited at that time. Meanwhile, motivated by the excellent results of DL in generic image classification, a significant amount of research on DL has been conducted in the last four years for HEP2IC and many have obtained significantly better results than the handcrafted methods proposed during the competitions. Although there is no specific review to cover these methods, a few surveys \citep{litjens2017survey,xing2017deep} reviewed some of these methods as a part of their broader scope. Table \ref{tab: existing surveys} shows a list of existing survey papers involving HEP2IC. Unlike the previous surveys, this paper mainly focuses on reviewing the DL based HEP2IC methods, organizes them in a usage based taxonomy, and discusses their core ideas and key achievements.

\begin{figure*}[t]
    \centering
    \includegraphics[width=0.95\textwidth]{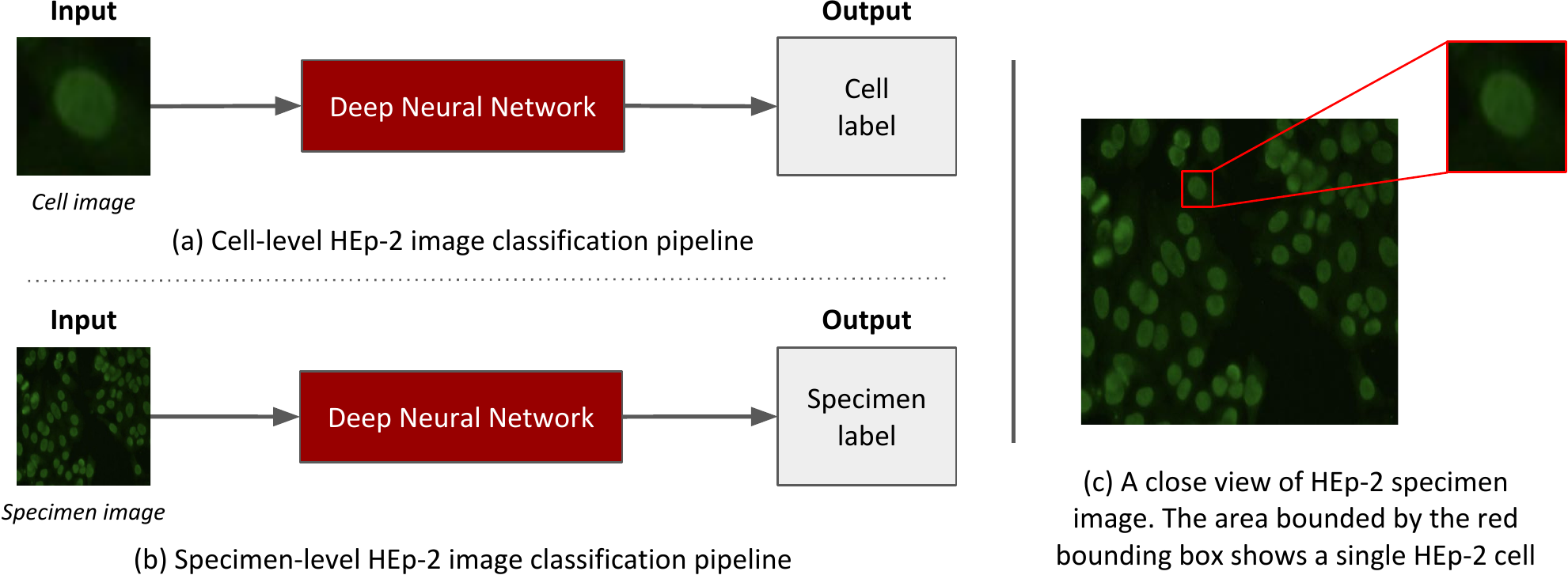}
    \caption{\label{fig:single vs specimen pipeline}(a) The pipeline of cell-level HEp-2 classification (CL-HEP2IC) methods which take a single HEp-2 cell image as input and output its label. (b) The pipeline of specimen-level HEp-2 classification (SL-HEP2IC) methods which take an HEp-2 specimen image as input and output its label. (c) An HEp-2 specimen image with zoom-in to a single cell. Better viewed in color.}
\end{figure*}

\textbf{Scope of this review.} There are two kinds of DL based HEP2IC methods available in the literature. Figure \ref{fig:single vs specimen pipeline} shows a high-level overview of both kinds of methods. The first kind takes a single HEp-2 cell image and predicts its class label. The second kind takes an HEp-2 specimen image and predicts its class label. Note that a specimen image contains many single cells. For simplicity, in the following discussions, we refer the single cell image based HEP2IC methods as the \textit{cell-level HEP2IC} (CL-HEP2IC) methods and the HEp-2 specimen based methods as the \textit{specimen-level HEP2IC} (SL-HEP2IC) methods. The issues of both CL-HEP2IC and SL-HEP2IC were introduced in ICPR 2013 and ICPR 2014 IIF image classification competitions, respectively. The CL-HEP2IC is a well-researched area and more than 20 papers have been published on this task till October 2019. On the other hand, the SL-HEP2IC is a relatively new area of research and the number of published papers is still low (i.e., less than 10 papers till October 2019). In light of the larger number of published papers, this review will primarily focus on reviewing the CL-HEP2IC methods. Meanwhile, the SL-HEP2IC methods will be discussed briefly. In short, Section \ref{sec: HEP2IC DL methods} will provide a review of both the CL-HEP2IC and SL-HEP2IC methods.

\section{Deep Learning for HEp-2 Cell Image Classification}
\label{sec: HEP2IC DL methods}

This section aims to review the existing state-of-the-art DL based methods for HEP2IC by highlighting the challenges and contributions from recent publications. At first, it gives an overview of the development of the DL based HEP2IC methods. Then, based on the classification task, the existing methods are grouped into two main categories, namely,  CL-HEP2IC and SL-HEP2IC. The key motivations, main ideas, and performance on the benchmark datasets of the methods from both categories are thoroughly discussed.

We begin with a brief introduction to DL. DL is a sub-area of machine learning that deals with artificial neural networks (ANN) inspired by the working principle of human brain. A typical kind of deep neural networks (DNNs) is the feed-forward ANNs with multiple hidden layers. DNNs are capable of learning from data in an automatic manner. Due to the powerful feature learning capability of DNN, it has been widely used in many computer vision applications, including HEP2IC. In HEP2IC, the spatial structure-preserving variants of DNN such as CNN are most commonly used. CNN has three basic components, convolution, pooling, and output layers. The convolution layers compute the output of locally connected neurons. The pooling layers perform downsampling of convolution layer output. And, an output layer computes the class probability scores. Many state-of-the-art CNN architectures have been proposed in recent years. For surveys on the CNN models, please refer to \citep{rawat2017deep} and \citep{zhao2019object}.

While it is common to use DNN as an automatic end-to-end classification framework in generic image classification such as classification of ImageNet \citep{krizhevsky2012imagenet}, some application domains such as medical image analysis have used the DNN for feature extraction only. This applied to HEp-2 cell image classification. As a result, in HEP2IC, DNN has been used as either feature extractor or an end-to-end classifier. When used as a feature extractor, for every image, a DNN generates a range of representations organized in a hierarchical manner which are used for classification. Specifically, in a CNN consisting of $n$ layers trained with supervised learning, the last layer is specified as a multi-class softmax function based on the number of target classes. To use the CNN as a feature extractor, the feature maps at the $(n-1)$th layer are usually extracted to form the image-level feature representation, upon which a separate classifier is trained. CNN and convolutional autoencoder (CAE) \citep{masci2011stacked} are the two spatial structure-preserving variants of DNNs that have been widely used as a feature extractor for HEP2IC. On the other hand, to use the CNN as an end-to-end classifier, the output class probability scores of the softmax function are regarded as the classification result. DNN generally requires a large number of training samples, and it will suffer from over-fitting otherwise \citep{srivastava2014dropout}. However, unlike generic image classification benchmark datasets, HEp-2 datasets are limited in sample size. Collection of large-scale datasets may not be impossible, but given the fact that the datasets must be compiled with accurate labelling by medical experts, it will be a challenging and time-consuming task. In order to mitigate this gap, most of the HEP2IC methods use data augmentation (DA) techniques such as rotation, flipping, and cropping. DA techniques are simple but effective in training the DNNs. Apart from DA, there are other strategies such as dropout \citep{srivastava2014dropout} and batch normalization \citep{ioffe2015batch}, which are often used with DNN to prevent overfitting.

\begin{figure*}[t]
    \centering
    \includegraphics[width=1\textwidth]{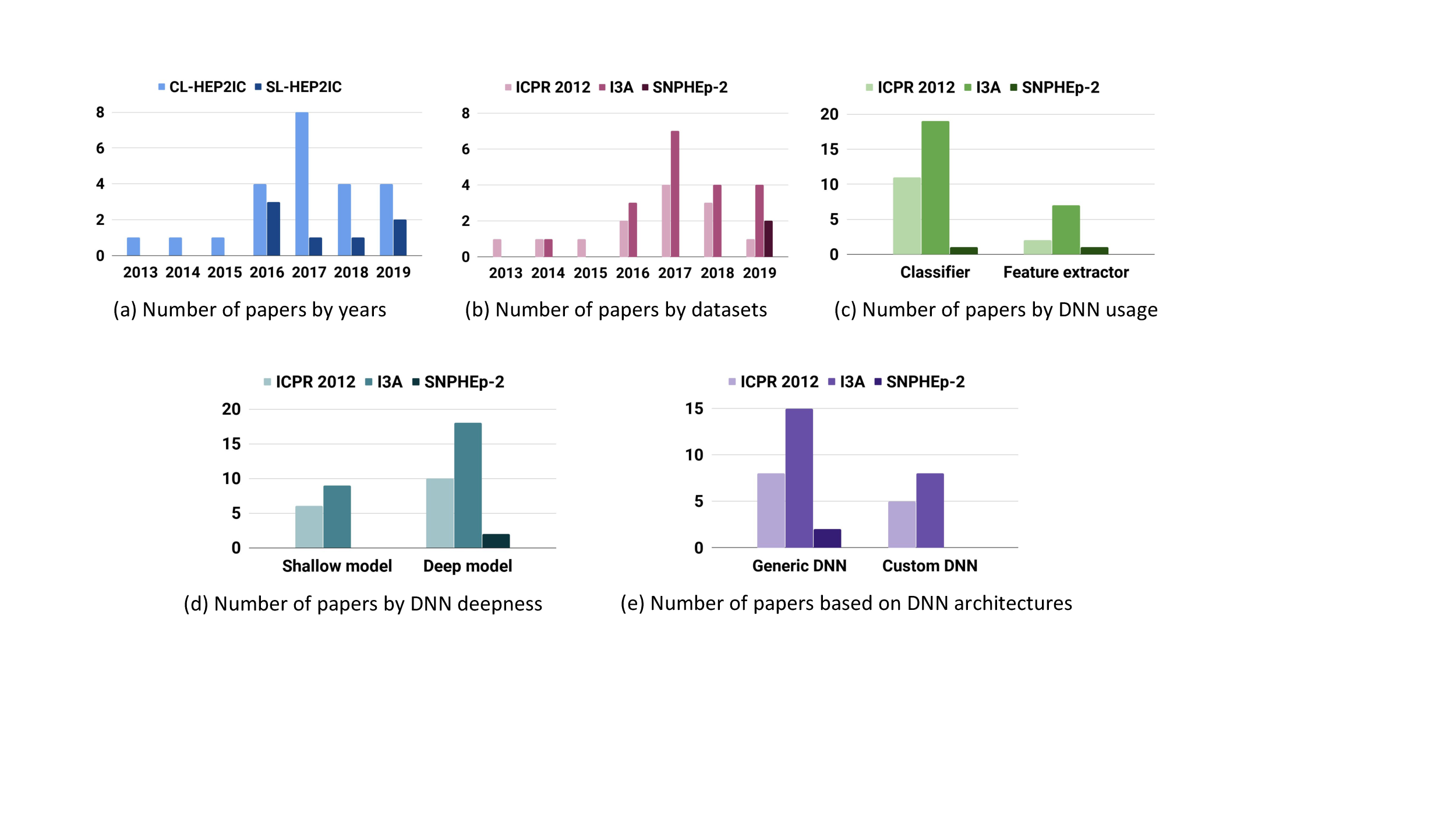}
    \caption{Breakdown of existing deep HEP2IC methods in terms of (a) number of publications per year, (b) yearwise number of publications per datasets, (c) number of publications based on the use of deep networks, (d) number of publications based on the deepness of deep models (models that have layers less than the VGG-16 \citep{simonyan2014very} models are considered as shallow models and models that have layers more or equal to VGG-16 are considered as deep models), (e) number of publications based on generic and customized networks used for HEP2IC. Better viewed in color.}%
    \label{fig:publications breakdown}
\end{figure*}

Figure \ref{fig:publications breakdown} shows the breakdown of the existing deep learning based HEP2IC literature. In particular, it scrutinizes the exiting literature as per the followings: (a) number of CL-HEP2IC and SL-HEP2IC papers by year, (b) number of papers per datasets, (c) number of papers by the type of DNN usage, (d) number of papers by the deepness of DNN architectures, and (e) number of papers by the type of DNN architecture (i.e., generic or customized). They are described in order as follows.

As shown in Figure 4(a), DL based HEP2IC has remained an active area since 2013. A large number of methods have been published, especially in the last four years. This could be attributed to the release of the indirect immunofluorescence images analysis (I3A) dataset \citep{foggia2014pattern} which has a relatively larger number of image samples (with approximately 14k samples) than the early HEp-2 dataset \citep{foggia2013benchmarking} and a top-ranked deep learning based method \citep{gao2014hep} in IIF image classification competition in 2014 \citep{foggia2014pattern}. Also, it is observed that in the early years, the research was conducted mainly for the CL-HEP2IC methods, however, SL-HEP2IC methods have also started to appear recently.

Figure 4(b) shows the growth of DL based CL-HEP2IC and SL-HEP2IC methods with respect to three most common HEp-2 datasets, namely, the International Conference on Pattern Recognition (ICPR) 2012 dataset, the Sullivan Nicolaides Pathology HEp-2 (SNPHEp-2) dataset and the I3A dataset. The ICPR 2012 dataset was one of the earliest datasets for HEP2IC. It was released in a public HEP2IC competition in 2012 and has only 1,455 cell images (extracted from 28 specimen images). Back then, training of high-performing DNNs with such a small dataset was challenging due to the problems such as over-fitting. The I3A dataset which is nine times larger than the earlier dataset was first released in 2013 and re-released in 2014 and 2016, and since then the HEP2IC research community started to take interest in classification methods with DL. Although developing larger HEp-2 datasets with sufficient classification difficulty is an expensive process, it will surely accelerate the DL based HEP2IC development.

Figure 4(c) segregates the existing DL based HEP2IC methods based on their usage of DNN. It demonstrates that most of the existing methods use DNN as the classifier to perform the classification of HEp-2 samples. At the same time, a small group of methods use DNN purely as a feature extractor. As shown in Figure 4(d), the majority of the existing HEP2IC methods employ deeper DNN architectures for better feature learning. The earliest HEP2IC methods use shallow DNNs such as LetNet-5 \citep{lecun1998gradient} and AlexNet \citep{krizhevsky2012imagenet}, but most of the recent methods employ deeper DNNs such as VGG-16 \citep{simonyan2014very} and ResNet-50 \citep{he2016deep}.

Figure 4(e) shows the type of DNN architectures used in the existing HEP2IC methods. As illustrated, most of the existing methods use generic DNNs for HEP2IC. Generic DNNs are the popular DNN architectures that have been originally designed for general image classification tasks but successfully applied to other application domains such as medical imaging. Examples of generic DNN include LeNet, VGG-Nets \citep{simonyan2014very} and ResNet \citep{he2016deep}, among others. However, some methods also use specially designed training schemes, new layers into generic DNNs, and fully customized DNN architectures. Figure 4(e) shows that such methods are only a few.

Motivated by the analysis above, in the following sections, the existing DL based HEP2IC methods are further discussed in details. For simplicity, a DNN usage-based taxonomy is defined below (also shown in Figure \ref{fig: taxonomy}) and used to review both the CL-HEP2IC and SL-HEP2IC methods.

\subsection{Cell-level HEp-2 image classification (CL-HEP2IC) methods}
\label{sec: CL-HEP2IC}
As aforementioned, considering the nature of DNN usage, there are two types of methods available, namely, (1) \textit{methods that use DNN as a feature extractor}, and (2) \textit{methods that use DNN as a classifier}. Table \ref{tab:cell-level methods} provides a comprehensive summary of both types of methods. The existing methods are organized based on their year of publication in an ascending order. Furthermore, for each method, the following key information are given: pre-processing techniques used, basic DNN architectures, dataset split and evaluation protocol used, performance on the benchmark datasets, and remarks. The table could be useful for the quick referencing of existing CL-HEP2IC methods. Sections \ref{sec: CL-HEP2IC-DNN-FE} and \ref{sec: CL-HEP2IC-DNN-CLFR} give a thorough discussion on these methods.

\begin{sidewaystable}[]
\scalebox{0.613}{%
\setlength{\tabcolsep}{0.8em} 
\begin{tabular}{lllllllll}\hline
\multirow{2}{*}{\textbf{Reference}} & \multicolumn{2}{l}{\textbf{Pre-processing}} & \multirow{2}{*}{\textbf{Classifier}}  & \multirow{2}{*}{\textbf{Dataset}} & \multirow{2}{*}{\textbf{\makecell[l]{Dataset Split/Used\\Evaluation Protocol}}} & \multicolumn{2}{l}{\textbf{Results (\%)}} &  \multirow{2}{*}{\textbf{Remarks}}\\ \cline{2-3}\cline{7-8}
& 	\makecell[l]{\textbf{EN}} &  \makecell[l]{\textbf{DA}} &   &    &  & \makecell[l]{\textbf{ACA}} &  \makecell[l]{\textbf{MCA}} & \\  \hline  
     \cite{foggia2013benchmarking} & CS & $\Cross$ &Custom CNN& ICPR 2012 & 5:0:5/Predefined set & 69.6 & 72.0& \makecell[l]{Process only single channel, illumination invariant, \\[0ex]  ignore background information, not robust against rotations.} \\
    \hline
     \multirow{2}{*}{\cite{gao2014hep}} & \multirow{2}{*}{CS}& \multirow{2}{*}{R}& \multirow{2}{*}{Custom CNN} & ICPR 2012 & 5:0:5/Predefined set & 69.9        & 72.0  & \multirow{2}{*}{\makecell[l]{Consider background information, rotation invariant, \\[-0ex] use of network ensemble during testing.}} \\ \cline{5-8}
     &&&& I3A & 6.4:1.6:2/Predefined set & $\Cross$ &96.76  & \\ \hline
     
     \cite{bayramoglu2015human} & CS, HE, MVN    & R, S, F, I, AOD   & Cuda-ConvNet  & ICPR 2012  & 5:0:5/Predefined set & 80.3 & $\Cross$& \makecell[l]{Rotation invariant, borrow new images, process only single channel,\\[-0ex] use too many training samples with DA, rely on MVN.} \\ \hline
     
     \cite{phan2016transfer}  & ZM & R & VGG-16+SVM  & ICPR 2012 & 5:0:5/Predefined set & 77.1        & $\Cross$ & \makecell[l]{Illumination invariant, rotation invariant, two-step classification,\\[-0ex] use ImageNet features, longer classifier training time.}\\ \hline
     
     \multirow{2}{*}{\cite{jia2016deep}} & \multirow{2}{*}{CS} & \multirow{2}{*}{M,F} & \multirow{2}{*}{Custom CNN} & ICPR 2012 & 8:0:2/Predefined set & 79.29       & 80.21         & \multirow{2}{*}{\makecell[l]{Does not use fine-tuning, class-balanced DA,\\[-0ex] over-fitting due to less samples.}} \\ \cline{5-8}
     &&&&I3A& 6.4:1.6:2/Predefined set & $\Cross$  & 98.26  & \\ \hline
     
     \cite{han2016hep} & $\Cross$ & $\Cross$ & Custom CNN & I3A & 5-fold cross validation &  $\Cross$ & 98.41 & \makecell[l]{Rotation invariant, no DA used, arbitrary input size,\\[-0ex] sensitive feature selection parameter, not end-to-end trainable.} \\ \hline
     
     \cite{li2016deep} & $\Cross$ & R, M, IDS & Modified LeNet-5 & I3A & Leave-one-specimen-out & $\Cross$ & 79.13$^{WS}$ & \makecell[l]{Deeper networks, training using related cross specimens,\\[-0ex] rotation invariant, use of more DA, tolerant to illumination changes.}\\ \hline
     
     \multirow{2}{*}{\cite{gao2017hep}} & \multirow{2}{*}{CS} & \multirow{2}{*}{R} & \multirow{2}{*}{Custom CNN} & ICPR 2012 & 5:0:5/Predefined set & 74.8 & 76.3 & \multirow{2}{*}{\makecell[l]{Consider background information, rotation invariant, use of network \\[-0ex]ensemble during testing, exhaustive use of rotations in training.}}\\ \cline{5-8}
     &&&& I3A & 6.4:1.6:2/Predefined set & $\Cross$  & 96.76  & \\ \hline
     
     \cite{liu2017hep} & $\Cross$ & R & DACN & ICPR 2012 & 5:0:5/Predefined set & 81.2 & $\Cross$ & \makecell[l]{Use encoder features for classification, no image enhancement is required,\\[-0ex] increase training parameters, not very suitable for transfer learning.} \\ \hline 
     
     \multirow{2}{*}{\cite{lei2017cross}} & \multirow{2}{*}{$\Cross$} & \multirow{2}{*}{$\Cross$} & \multirow{2}{*}{ResNet-50} & ICPR 2012 & 5:0:5/Predefined set & 95.63 & $\Cross$ & \multirow{2}{*}{\makecell[l]{No data preprocessing is required, trainable with smaller datasets,\\[-0ex] suffer from gradient vanishing problem and overfitting.}}\\ \cline{5-8}
     &&&& I3A & 6.4:1.6:2/Predefined set & $\Cross$  & 96.87  &\\ \hline
     
    \multirow{2}{*}{\cite{lu2017hep}} & \multirow{2}{*}{CS} & \multirow{2}{*}{$\Cross$} & \multirow{2}{*}{VGG-16+SVM} & ICPR 2012 & 5:0:5/Predefined set &  $\Cross$ & 92.2 &\multirow{2}{*}{\makecell[l]{No DA is required,\\[-0ex] no clear details of fine-tuning the CNN is given.}}\\ \cline{5-7}
    &&&& I3A & 8:2/Predefined set & $\Cross$  & 98.0  &\\ \hline
    
    \cite{li2017joint} & CS & $\Cross$ & MFC-ELM-CNN & I3A & Not specified &  81.1 &$\Cross$ & \makecell[l]{Good feature generalization capacity,\\[-0ex] computationally expensive, large parameter size.} \\ \hline
    
    \cite{li2017deep} & $\Cross$ & R,S & DRI-Net & I3A & 6.4:1.6:2/Predefined set & 98.49 & 98.37 & \makecell[l]{Multi-scale feature extraction, auxiliary supervision \\[-0ex]for efficient classification, suffer from over-fitting, longer training time.} \\ \hline
    
    \cite{rodrigues2017exploitingnpreprocessing} & CS, AS, HE & R & \multirow{2}{*}{\makecell[l]{LeNet-5, \\ AlexNet,\\ and GoogleNet}} & I3A & 6.4:1.6:2/Predefined set & $\Cross$ & 98.17 & \makecell[l]{Use effective pre-processing techniques, invariant to \\[0ex]illumination and rotation, use too many pre-processing techniques.} \\ \cline{1-3} \cline{5-9} 
    
    \cite{rodrigues2017hep} & CS, AS & $\Cross$ &  & I3A & 10-fold cross val. & $\Cross$& 95.53& \makecell[l]{No DA is used, invariant to illumination,\\[0ex] tolerant to rotation.}\\ \hline
    
    \multirow{2}{*}{\cite{lei2018deeply}} & \multirow{2}{*}{$\Cross$} & \multirow{2}{*}{$\Cross$} & \multirow{2}{*}{ResNet-50} & ICPR 2012 & 5:0:5/Predefined set & 97.14 & $\Cross$ & \multirow{2}{*}{\makecell[l]{No data preprocessing is required, reduce gradient vanishing problem,\\[-0ex] increase the network parameters.}}\\ \cline{5-8}
    &&&& I3A & 8:0:2/Predefined set & 98.42& $\Cross$  &\\ \hline
    
    \multirow{2}{*}{\cite{li2018hep}} & \multirow{2}{*}{$\Cross$} & \multirow{2}{*}{R, S} & \multirow{2}{*}{HEp-Net} & ICPR 2012 & 5:0:5/Predefined set & 78.9 & $\Cross$ & \multirow{2}{*}{\makecell[l]{Fewer parameters to train, suitable for training with smaller datasets,\\[-0ex] may suffer from overfitting, tolerant to illumination changes.}}\\ \cline{5-8}
    &&&& I3A & 6.4:1.6:2/Predefined set & 98.96 & 98.50  &\\ \hline
    
    \multirow{2}{*}{\cite{shen2018deep}} & \multirow{2}{*}{CS} & \multirow{2}{*}{R, F, C} & \multirow{2}{*}{DCR-Net} & ICPR 2012 & 5:0:5/Predefined set & 80.8 & $\Cross$ & \multirow{2}{*}{\makecell[l]{Deeper layers, fewer parameters to train, multi-scale feature extraction,\\[-0ex] too many cross connections between layers, use of more DA.}} \\ \cline{5-8}
    &&&& I3A & 8:0:2/Predefined set & 98.82 & 98.62  &\\ \hline
    
    \cite{ebrahim2018performance} & ZM & R,C & Custom CNN & I3A & 7.5:2.5/Predefined set & 98.29 & $\Cross$ & \makecell[l]{Invariant to illumination and rotation, introduce online DA,\\[-0ex] suffer from noise, over-fitting, too many DA.} \\ \hline
    
    \cite{majtner2019effectiveness} & CS & R, GAN & \makecell[l]{VGG-16, \\[-0ex] GoogleNet,\\[-0ex] and Inception-v3} & I3A & 7:1:2/Predefined set & 98.60 & 98.71 & \makecell[l]{Use synthetic image by GAN in training, systematic DA study,\\[-0ex] use of synthetic image deteriorates classification performance.} \\ \hline
    
    \cite{nguyen2019biomedical} & $\Cross$ & $\Cross$ & Ensemble Net & ICPR 2012 & 8:0:2/Predefined set & 94.98 & $\Cross$ & \makecell[l]{Simple transfer learning idea, effective feature extraction, \\[-0ex] inefficient due to large feature dimension, used pre-trained features.} \\ \hline
    
    \multirow{2}{*}{\cite{vununu2019dynamic}} & \multirow{2}{*}{DWT} & \multirow{2}{*}{R} & \multirow{2}{*}{\makecell[l]{Dynamic\\[-0ex]Learning Net}} & SNPHEp-2 & 5:0:5/5-fold cross. val. & 98.27 & $\Cross$ & \multirow{2}{*}{\makecell[l]{Handle homogeneous image classes, simple and effective network,\\[-0ex] number of network parameter increased}.} \\ \cline{5-8}
    &&&& I3A & Not specified & $\Cross$ & 98.89  &\\ \hline
    
    \cite{cascio2019deep} & CS & R & \makecell[l]{AlexNet+SVM\\[-0ex]+k-NN} & I3A & Leave-one-specimen-out & 81.93$^{WS}$ & 82.16$^{WS}$ & \makecell[l]{Simple transfer learning, two-level classification,\\[-0ex] use of generic image features, complex training of binary classifiers.} \\ \hline
    
    \multirow{2}{*}{\cite{vununu2019deep}} & \multirow{2}{*}{\makecell[l]{Gradient\\[-0ex] image}} & \multirow{2}{*}{$\Cross$} & \multirow{2}{*}{\makecell[l]{Dual DCAE+\\[-0ex] Shallow ANN}} & SNPHEp-2 & 5:0:5/5-fold cross val. & 98.27 & $\Cross$ & \multirow{2}{*}{\makecell[l]{Capture local and global image information, compact feature \\[-0ex]representation, two-step classification framework, two-stream network.}} \\ \cline{5-8}
    &&&& I3A & 8:0:2/Predefined set & 98.66 & $\Cross$  &\\ \hline
    
\end{tabular}}
\caption{\scriptsize Summary of the existing methods for cell-level HEp-2 image classification. The widely accepted performance criterion for ICPR 2012 and SNPHEp-2 datasets is ACA (average classification accuracy), and for I3A dataset is MCA (mean classification accuracy) (please refer to Section \protect\ref{sec: datasets} for details). All reported results are quoted from original papers. The results are produced by the respective authors using publicly accessible part of the datasets and customized evaluation protocols (please refer to Section \protect\ref{sec: datasets} for details). Data split ratio is reported in training:validation:test format. EN=Image enhancement method; CS = Contrast stretching; HE = Histogram Equalization; ZM = Zero mean; MVN = Zero mean and unit variance normalization; DA = Data augmentation; R = Rotation; S = Shifting; F = Flipping; I = Intensity variations; C = Cropping; AOD = Adoption from other datasets; IDS = Import from related datasets; WS = Specimen-level classification with leave-one-specimen-out protocol.}
\label{tab:cell-level methods}
\end{sidewaystable}

\subsubsection{CL-HEP2IC methods that use DNN as a feature extractor}
\label{sec: CL-HEP2IC-DNN-FE}
Feature extraction is a common step in many medical imaging applications including HEP2IC. It is often used in the handcrafted feature based HEP2IC methods, e.g., \citep{nosaka2014hep}, \citep{liu2014hep}, and \citep{shen2014hep}. As the name suggests, the feature extraction step extracts necessary features from the input image to perform the classification, i.e., labelling. The state-of-the-art classifiers such as support vector machines (SVM) \citep{cortes1995support} and k-nearest neighbours (k-NN) \citep{cover1967nearest} are the popular choices in the literature for classification. While the feature extraction and classification steps are treated separately in the handcrafted feature based methods, the DNN based methods combine them as one integral part. As mentioned previously, a DNN could be considered as a hierarchical feature extractor, which is organized in a bottom-up fashion. The hierarchies are usually defined by the layers and each layer takes the input data, performs some operations on it, and then delivers the result as output. The output of each layer could also be treated as the features and can be used for training a separate classifier. Overall, depending on the training status of DNN, there are two types of feature extraction in the existing HEP2IC methods, namely, \textit{feature extraction from a pre-trained DNN model} and \textit{feature extraction from a fine-tuned DNN model}. Both types are discussed below.

\noindent\textbf{Feature extraction from a pre-trained DNN model.} Methods of this kind use the basic form of transfer learning, where features from the DNN trained on a large-scale dataset such as ImageNet \citep{krizhevsky2012imagenet} are extracted \citep{sharif2014cnn} and used for the training of a separate classifier. The methods proposed by \citep{lu2017hep} and \citep{phan2016transfer} are the examples of this kind. Both of them use CNN features trained on the ImageNet dataset. The method proposed by \citep{lu2017hep} is relatively straightforward. They directly replaced the handcrafted feature descriptors such as scale-invariant feature transform (SIFT) \citep{lowe1999object} used in the traditional image classification pipeline \citep{yang2007evaluating} with the CNN features. Specifically, the convolution features (i.e., \texttt{conv\_5}) of a VGG-16 network are extracted and a multi-class SVM with a radial basis function (RBF) kernel \citep{scholkopf1997comparing} is trained with them. Even though the ImageNet data are considerably different from the HEp-2 samples, this method achieved a very good performance on both the ICPR 2012 and the I3A datasets.

On the other hand, \citep{phan2016transfer} also used ImageNet pre-trained features but adopted a more complex classification framework. Unlike the above method, they proposed an intensity-aware two-step classification framework with late feature fusion. Furthermore, a feature selection approach \citep{peng2005feature} is applied to extract the most discriminative features for the training of class-specific SVMs at different stages of the framework. However, their performance is not as high as that of \citep{lu2017hep}'s method.

A more recent work proposed by \citep{cascio2019deep} also uses a two-step classification framework. Given the CNN features, the proposed framework firstly passes them into a set of class-aware binary SVM's to generate a compact and discriminative feature representation. Next, a k-NN classifier is used to classify the compact cell features. However, this method produces lower performance than the above two methods \citep{lu2017hep,phan2016transfer} and requires segmentation masks to obtain a discriminative feature set. Moreover, the idea of training class-aware SVM's may not scale well for datasets with a large number of classes.

Since there exists a domain gap between the ImageNet and HEp-2 data, a few methods in the literature have used fine-tuned CNN models \citep{tajbakhsh2016convolutional} for better HEP2IC. The following paragraphs will focus on the methods that use fine-tuned DNN extracted features.

\noindent\textbf{Feature extraction from a fine-tuned DNN model.} Fine-tuning pre-trained DNN models to target datasets generally improves the discriminative capacity of the features, hence, leads to better classification performance. Interested readers are referred to a recent work on this topic \citep{tajbakhsh2016convolutional} for details on fine-tuning. It is worth mentioning that the process of feature extraction from the fine-tuned CNN is similar to that from the pre-trained CNNs. Among the existing methods of this kind, \citep{han2016hep} used a fine-tuned CaffeNet \citep{jia2014caffe} for feature extraction. While it is common for the CNNs to resize the input image in order to have a fixed-sized input (in other words, size normalized input), the proposed method developed a new pooling strategy called `\textit{K}-spatial pooling' to support the HEp-2 cell images with arbitrary sizes. The idea behind \textit{K}-spatial pooling is to leverage the frequency of neural activation patterns in feature pooling. Given a set of convolutional activation maps (also known as feature maps), the proposed pooling strategy finds the \textit{K} larger activation values in a defined region from each of feature maps and performs a mean operation on them. Suppose we have $N$ feature maps and $M$ pre-defined regions in each of the feature maps, the \textit{K}-spatial pooling output will be $\mathbf{z} = [\Psi(F_1^M), \Psi(F_2^M), ..., \Psi(F_N^M)]^T$ where $\Psi(\cdot)$ calculates the mean values of \textit{K} larger activations from feature maps $F$. The feature vector $\mathbf{z}$ can be further used for the training of common classifiers such as SVM. The experimental results in \citep{han2016hep} show that the proposed pooling method is able to extract more discriminative and rotation invariant features from the convolutional activation maps than the traditional max-pooling. However, the parameter \textit{K} is sensitive and determined by empirical evaluations, which is a time-consuming task. To improve this situation, instead of using a pre-calculated value of \textit{K} during the feature extraction, automatically learning it during the fine-tuning of pre-trained CNN model could be a more efficient way.

While the above methods choose CNN as the feature extractor, \citep{vununu2019deep} used the CAE to extract features for CL-HEP2IC. Specifically, they trained two distinct CAEs with the regular (i.e., RGB) and gradient images. While the regular image based CAE learns the geometrical information of HEp-2 cells, the gradient image based CAE learns the local intensity changes in HEp-2 cells. The CAE is based on the VGG-Net architecture. Unlike CNN, the latent space features between the encoder and decoder of CAE are extracted. The decoder part of CAE uses the latent space features to reconstruct the input, hence, it presumed to carry rich information of the HEp-2 cells. The latent space features from the above networks are then combined and classified using a simple neural network based classifier. The joint utilization of cells' geometrical and local intensity information significantly improves the classification performance, as demonstrated by the state-of-the-art results on both the SNPHEp-2 and the I3A datasets.

\noindent\textbf{Summary of discussion.} Feature extraction based methods use image features that have been automatically learned by the DNN to perform the CL-HEP2IC. Both the pre-trained and fine-tuned DNN models have been used for feature extraction. However, due to the domain gap between the data used for pre-training and the HEp-2 cell images, the performance of pre-trained model based methods are usually lower than that of the fine-tuned model based methods. Nevertheless, efforts have been made to improve the performance of the former, and they include  multi-step and class-specific classification frameworks and discriminative feature selection \citep{cascio2019deep,phan2016transfer}. On the other hand, features extracted from fine-tuned models are more powerful and do not require multi-step frameworks or explicit feature selection. The early feature extraction based methods are based on CNN, whereas the recent methods also use CAE features.

The common classifiers used in feature extraction based methods are SVM and k-NN. However, for classification tasks, DNNs are usually trained using the softmax classifier \citep{litjens2017survey} and it is possible to directly use it for CL-HEP2IC. In the literature, most of the DNN based HEP2IC methods directly use softmax predictions and avoid training a separate classifier such as SVM, as discussed in the following part.

\subsubsection{Cell-level HEp-2 (CL-HEP2IC) methods that use DNN as a classifier}
\label{sec: CL-HEP2IC-DNN-CLFR}
The methods of this kind treat the feature extraction and classification steps jointly. Given the image, features are automatically extracted, and then classification is performed using the extracted features. Based on the use of network architectures and their training and fine-tuning strategies, the DNN-as-a-classifier based CL-HEP2IC-methods are decomposed into three groups, namely, (1) pure generic DNN based methods, (2) generic DNN based methods with partial changes in layers or training schemes, and (3) customized DNN based methods. They are thoroughly discussed below.

\textbf{Pure generic DNN based methods.} Generic DNN based methods use the popular DNN architectures to perform CL-HEP2IC. The methods proposed in \citep{bayramoglu2015human}, \citep{lei2017cross}, \citep{majtner2019effectiveness}, \citep{rodrigues2017exploitingnpreprocessing}, and \citep{rodrigues2017hep} are of this kind. Since the generic CNNs are primarily designed for the general image classification tasks such as ImageNet classification \citep{krizhevsky2012imagenet}, pre-processing techniques such as image enhancement and DA (i.e., data augmentation) are carefully considered in most of these papers to achieve high CL-HEP2IC performance. A list of pre-processing techniques used in these methods is given in Table \ref{tab:cell-level methods}. 

One of the early and successful DL based HEP2IC methods proposed by \citep{bayramoglu2015human} uses the AlexNet CNN architecture. Since AlexNet was originally proposed for the ImageNet classification, the authors have performed a comprehensive study on various pre-processing techniques and used the best-performing techniques to train the network. They have managed to achieve the state-of-the-art performance on the ICPR 2012 dataset. One of the interesting strategies they have followed was the adoption of samples from the I3A dataset and the SNPHEp-2 dataset for the training. They have showed that their training scheme is very useful for achieving good performance on the smaller HEp-2 datasets.

Another two studies conducted by \citep{rodrigues2017exploitingnpreprocessing,rodrigues2017hep} also performed a comprehensive review on various pre-processing techniques for the training of generic CNNs. They considered three well known CNN architectures, namely, AlexNet, GoogleNet \citep{szegedy2015going}, and LetNet-5 \citep{lecun1998gradient}. In both of their studies, they demonstrate that without any pre-processing used, GoogleNet outperforms the other architectures on the I3A dataset. This indicates that the deeper CNN model is more robust against illumination changes and is able to generate more robust and discriminative features than the shallow models. In a most recent extention of their work \citep{rodrigues2020comparing}, the authors further evaluated VGG-16 \citep{simonyan2014very}, Inception-V3 \citep{szegedy2016rethinking}, and ResNet-50 \citep{he2016deep} models on CL-HEP2IC. Alongside, they also performed an evaluation on the tree of Parzen estimators algorithm \citep{bergstra2011algorithms} for CNN hyper-parameters optimization during fine-tuning. Their latest work confirms that Inception-V3 model is capable of giving better classification performance (i.e., under five-fold cross validation) than that of other CNN models considered in the study without any data pre-processing. Meanwhile, \citep{lei2017cross} proposed an HEP2IC method based on the ResNet architecture \citep{he2016deep}. They used the ICPR 2012 dataset without any pre-processing to train the ResNet-50 CNN model. Their method achieves the state-of-the-art classification performance on the I3A dataset. Unlike most of the generic CNN model based CL-HEP2IC methods, \citep{lei2017cross}'s method first pre-trains the CNN with the smaller ICPR 2012 datasets, and then applies fine-tuning to the larger dataset, i.e., the I3A dataset.  Both studies in \citep{lei2017cross} and \citep{rodrigues2017exploitingnpreprocessing, rodrigues2017hep} suggest that the deep CNNs are capable of generating more robust features than the shallow CNNs.

In \citep{rodrigues2017exploitingnpreprocessing,rodrigues2017hep,rodrigues2020comparing} and other similar studies, the authors only considered employing a single CNN stream for HEP2IC. Unlike them, \citep{nguyen2019biomedical} studied the network ensemble for HEP2IC. However, their study is only limited to pre-trained models. The motivation behind their study is CNN's over-fitting tendency with smaller datasets, e.g., ICPR 2012. To avoid the fine-tuning of CNN features, they extracted a wide range of pre-trained features. Specifically, six types of ResNet and GoogleNet based ImageNet pre-trained models are used for the feature extraction. Since it is computationally expensive to combine all of the extracted features, the features are averaged and then used for the softmax classification. The findings of \citep{nguyen2019biomedical}'s method are interesting. They showed that combining the features extracted from various models further enhances the classification performance. At the same time, we presume that their method could also benefit from fine-tuning. In addition, instead of simply averaging the features, a weighted feature fusion mechanism could be adopted to obtain a more effective image representation.

Another interesting work in this group was proposed by \citep{majtner2019effectiveness}. They propose to perform generative adversarial network (GAN) \citep{radford2015unsupervised} based DA. In existing HEP2IC methods, DA is performed using simple image transformations such as rotation and flipping. Unlike the existing methods, Majtner et al. proposed to generate synthetic HEp-2 cell images by the GAN. Different from rotation and flipping, GAN produces images with different compositions than the original images. 
The proposed method by \citep{majtner2019effectiveness} trained recent CNN architectures such as GoogleNet with GAN-produced images to investigate their effectiveness. The findings are interesting. It shows that the traditional DA methods, e.g., rotation, are more effective than GAN based DA. Furthermore, the authors mentioned that the GAN is not robust against the large intra-class variation of data and as a result produces HEp-2 synthetic images of poor quality. The authors also emphasized on continuing the research of GAN to solve the problems of datasets with a limited number of annotated samples.

\textbf{Generic DNN based methods with partial changes in layers or training schemes.} The methods in this group make minor changes in the generic network architectures or use special training schemes for robust feature learning from HEp-2 cell images. Two existing methods \citep{lei2018deeply, li2016deep} fall under this category. The summary of both methods is given in Table \ref{tab:cell-level methods}.

The first method is proposed by \citep{li2016deep}. They add two additional covolutional layers with $1\times 1$ filters before the classification layers (i.e., fully connected and softmax) of LeNet-5 \citep{lecun1998gradient} CNN architecture. Their motivation of using these additional layers is to increase the number of feature channels for the classification layers. The network is trained from scratch with a relatively large dataset compiled from both the I3A Task-1 and Task-2 datasets (the details of Task-1 and Task-2 datasets are further discussed in Section \ref{sec: datasets}). Note that this method primarily deals with the specimen classification and uses SL-HEP2IC based evaluation criterion. However, due to its individual image based classification strategy and the performance on I3A Task-1 dataset, we include this method under CL-HEP2IC. Their experimental results show that the proposed architecture outperforms the baseline LeNet-5 by a significant margin. 

The second method is proposed by \citep{lei2018deeply}. It enhances the training and classification performance of the ResNet-50 architecture by combining the early layer predictions into the final classification layer. In particular, a parametric bridging mechanism is proposed to connect the early layers to the final layer. The experimental result shows that combining the classification decisions of early layers into the final layer has a positive impact on the final classification performance. This method achieved state-of-the-art result in both the ICPR 2012 and the I3A datasets. The major drawback of this method is that it significantly (almost double of the baseline) increases the number of network parameters.

\textbf{Customized DNN based methods.} A large number of customized DNN based methods \citep{ebrahim2018performance,foggia2013benchmarking,gao2017hep,gao2014hep,jia2016deep,li2017joint,li2018hep,li2017deep,liu2017hep,shen2018deep,vununu2019dynamic} have been proposed in recent years, especially in the last four years. While only a few of them make partial customizations to the generic DNN architectures, most of the methods use fully customized DNN architectures. Both are discussed below.

\begin{figure}[t]
    \centering
    \includegraphics[width=0.85\textwidth]{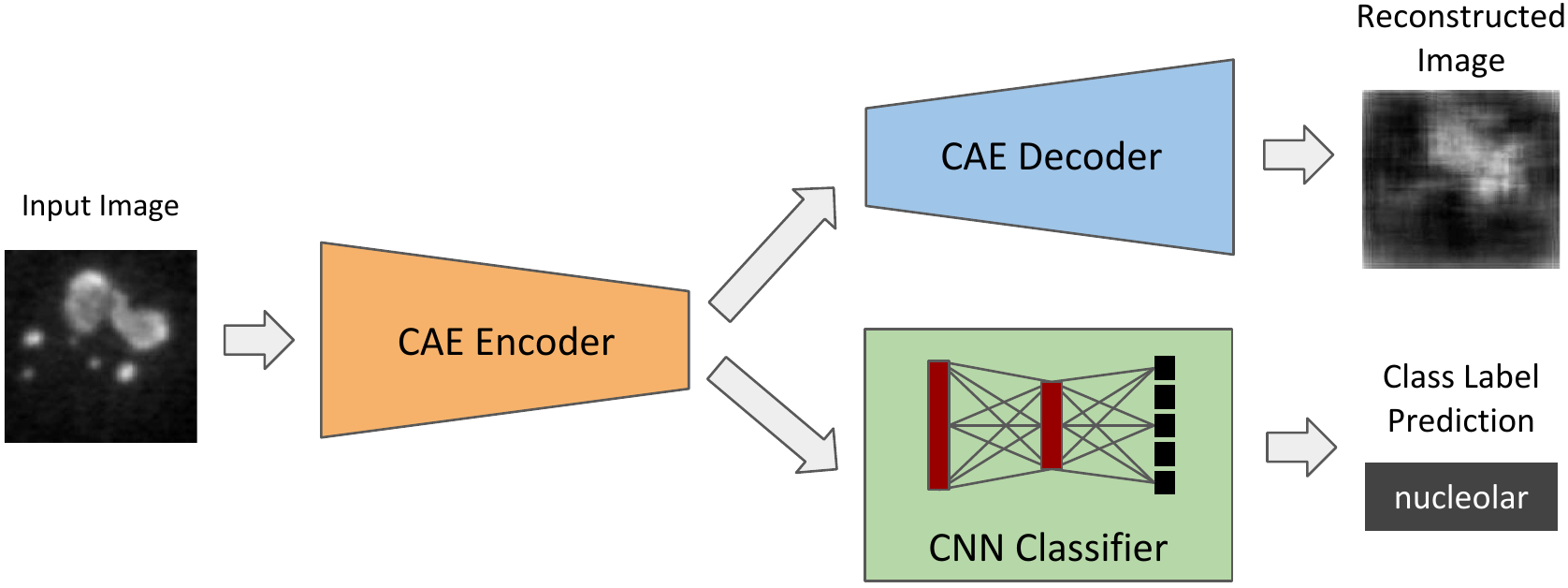}
    \caption{Deep autoencoding-classification network (DACN) proposed by \citep{liu2017hep}.}
    \label{fig:dacn}
\end{figure}

\begin{figure}[h]
    \centering
    \includegraphics[width=0.85\textwidth]{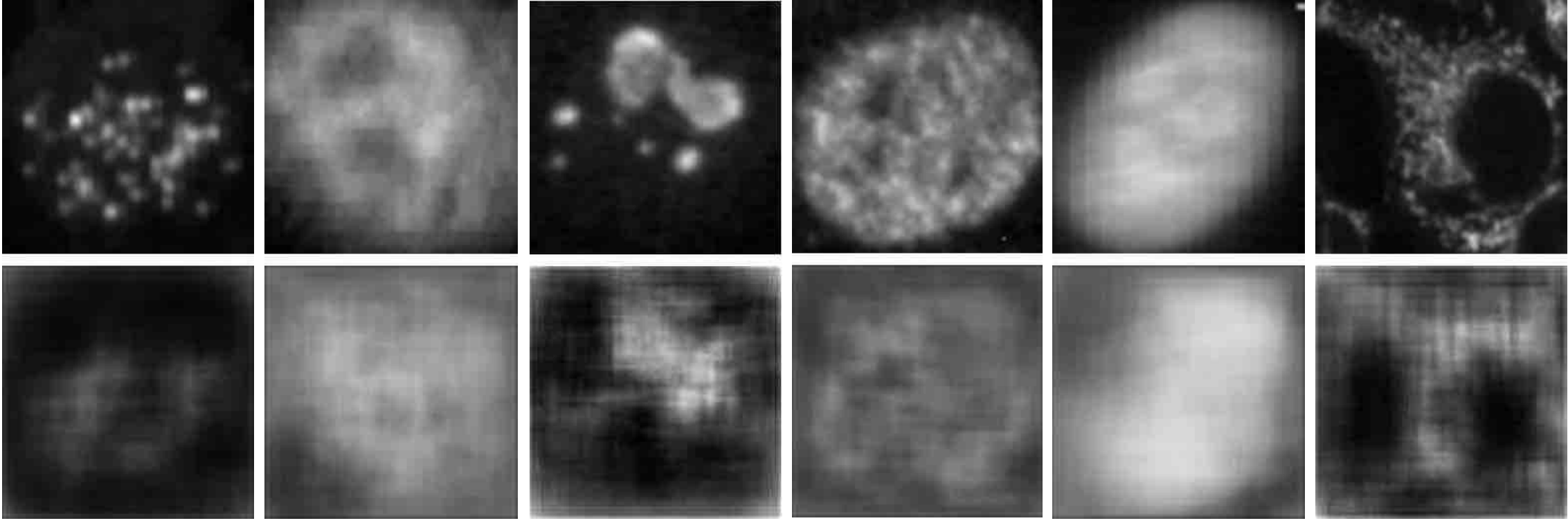}
    \caption{Input and reconstructed images by the CAE proposed by \citep{liu2017hep}. Top and bottom rows show the input and reconstructed images, respectively. Image courtesy of \citep{liu2017hep}.}
    \label{fig:dacn-maps}
\end{figure}

\textit{Methods that made partial customizations to the generic DNN models.} \citep{liu2017hep}'s method is one of the very few to make partial modifications to the CAE for HEP2IC. They combined an unsupervised CAE with supervised CNN classification branch that takes raw image data and predicts their class labels. It is a multi-task network optimized over the classification and reconstruction losses. Figure \ref{fig:dacn} shows the network proposed by \citep{liu2017hep}. Let us denote $F_a$ as the unsupervised CAE auto-encoder model, $F_c$ as the supervised CNN classification model, and $W_a$ and $W_c$ as their learnable parameters, respectively. The network is optimized by the loss function $\mathcal{L}$ as follows:
\begin{equation}
\mathcal{L}_T = \lambda \mathcal{L}_a(W_a) + \mathcal{L}_c(W_{c})
\end{equation}
where $\mathcal{L}_T$ is the total loss, $\mathcal{L}_a$ is the reconstruction loss, $\mathcal{L}_c$ is the classification loss, and $\lambda$ is the trade-off hyper-parameter. The idea of using an encoder where hidden layer features are able to reconstruct the output helps improve the feature representation in the classification layer, as demonstrated by the state-of-the-art classification result achieved by the proposed method on the ICPR 2012 dataset. Figure \ref{fig:dacn-maps} shows some example inputs and reconstructed images produced by the CAE in \citep{liu2017hep}. Meanwhile, based on our observation, there could be some issues by using CAE for classification, such as, (1) compared with CNN, it increases the number of training parameters in the network and (2) it strives to capture the underlying manifold of training data, which might become an issue if transfer learning to a slightly different dataset is needed.

\citep{li2017joint} proposed to combine the CNN with fully connected networks and extreme learning machines (ELMs) \citep{huang2006extreme}. In that method, the authors replaced the CNN softmax layer with ELMs. Furthermore, they have used multiple fully connected layers to combine the features from the final convolutional layer of a CNN into the ELMs (referred as multi-form feature extraction by the authors). The motivation behind this is to take advantage of the CNN's powerful features. The obtained features are then used to train the ELMs for label prediction. Their method has a better feature generalization capacity than the CNN based methods. As shown in an experiment, the proposed network is pre-trained using a grading hepatocellular carcinoma image dataset \citep{li2017joint} and achieves a decent ($\sim$81\%) classification performance on the I3A dataset. However, this method has more trainable parameters than the CNN, and some of the layers are designed carefully to guarantee the best feature representation for the task.

\begin{figure}[h]
    \centering
    \includegraphics[width=0.75\textwidth]{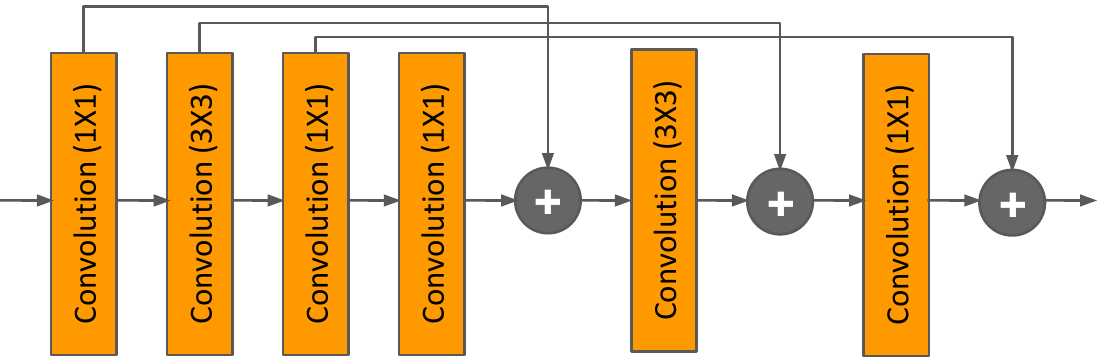}
    \caption{Deep cross-residual (DCR) module with three cross connections proposed by \citep{shen2018deep}.}
    \label{fig:deep residual block}
\end{figure}

\citep{shen2018deep} proposed modified residual block to minimize the vanishing gradient problem in residual networks \citep{he2016deep} for CL-HEP2IC. They modified the original residual block by adding cross-residual shortcuts between its various components, i.e., layers. The modified block has two major advantages over the original block, namely, (1) improved feature representation due to cross-layer information sharing and multi-scale feature extraction and (2) computational savings, i.e., the modified block being able to increase the network depth by two times with 26.9\% less parameters than that of the original block. Figure \ref{fig:deep residual block} shows the deep cross-residual (DCR) module proposed by \citep{shen2018deep}.

\citep{li2017deep} proposed deep residual inception network (DRI-net) by modifying the basic convolution blocks in the residual module of residual network \citep{he2016deep}. DRI-net integrates the key advantages of two high-performing CNN architectures, namely, Inception-net \citep{szegedy2016rethinking} and ResNet. Specifically, it combines the multi-scale feature extraction approach of Inception-net and efficient network optimization approach of ResNet. Furthermore, it leverages auxiliary classifier decisions taken with the features from the early, middle, and end layers to improve the classification decisions. They also modified the design of the original inception module for better feature representation and for overcoming the vanishing gradient problem when the number of layers in network increases. Precisely, batch normalization and parametric rectified linear unit (PReLU) \citep{he2015delving} are used before all the convolution layers, and two new identity shortcuts are added to the network. However, it is observed from the reported experimental results that the DRI-net is not suitable to train with smaller HEp-2 cell datasets and often requires longer training duration. The multi-scale convolutional component (MCC) module in DRI-net is primarily responsible for this issue.

\begin{figure}[h]
    \centering
    \includegraphics[width=0.85\textwidth]{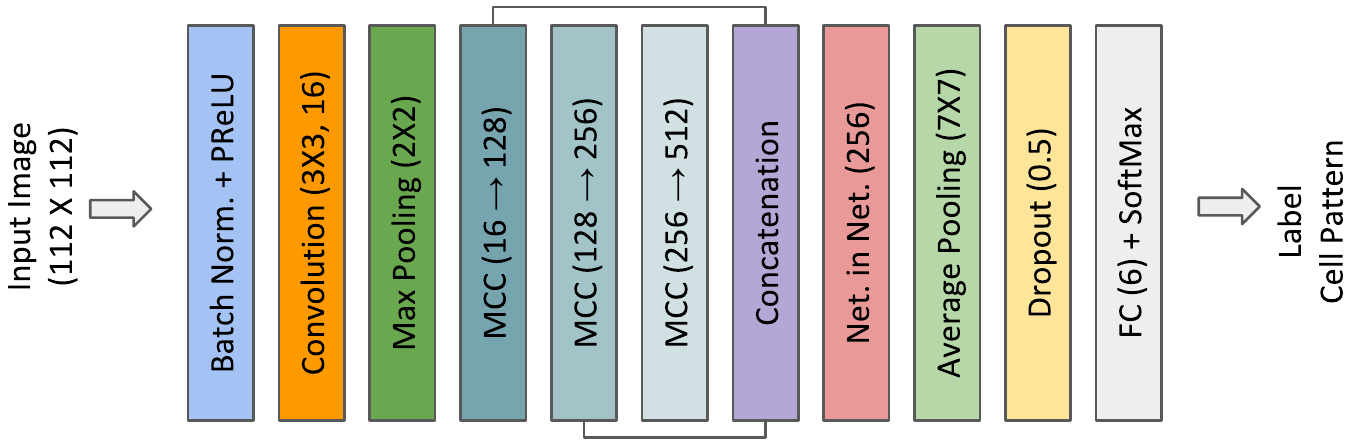}
    \caption{Architecture of the HEp-Net proposed by \citep{li2018hep}. The network is comprised of batch normalization, convolution, max-pooling, MCC block, network-in-network block, average pooling, and dropout operations. A joint feature representation of all the MCC blocks are used during training.}
    \label{fig:hep-2 architecture}
\end{figure}

In a more recent work \citep{li2018hep} of \citep{li2017deep}, the authors further improved the MCC module by (1) reducing the size of its convolution layers, (2) expanding the convolution from two to three scales, and (3) introducing a new global shortcut as a replacement of two identity shortcuts. The improved version of the MCC module is capable of extracting rich features at a higher speed, which significantly reduces the network training time. A new deep architecture named `HEp-Net' based on this improved MCC module is proposed in \citep{li2018hep}. The HEp-Net is very lightweight, i.e., only 7\% of the DRI-net size (the number of parameters of DRI-net is about 1,985M), but is capable of learning rich features from smaller HEp-2 datasets. Figure \ref{fig:hep-2 architecture} shows the architecture of the HEp-Net.

\textit{Methods that rely on fully customized DNN models.} Malon et al. (2012)\footnote{The details of method proposed by Malon et al. (2012) is available at \url{https://mivia-web.diem.unisa.it/hep2contest/results.html}} first attempted to use CNN for CL-HEP2IC in the ICPR 2012 HEp-2 Image Classification Competition \citep{foggia2013benchmarking}. They have proposed a simple CNN to perform CL-HEP2IC on the ICPR 2012 dataset. To deal with the illumination variations in the cell images, they have used absolute value rectification and subtractive spatial normalization in the CNN. Although the proposed method did not manage to achieve superior performance (i.e., 6th place), it outperformed many specially designed handcrafted methods \citep{foggia2013benchmarking} and later inspired many researchers to use the CNN for HEP2IC. Issues related to Malon et al. (2012)'s method are the neglect of the background surrounding the cell contours which is later proved to be important for separating similar cell classes \citep{gao2017hep} and the insufficient consideration to the common problems such as cell rotations.

\citep{gao2017hep,gao2014hep} first proposed the successful DL based method for HEP2IC. They used a shallow CNN model consisting of three convolution layers, three pooling layers, and a fully connected layer. The CNN was trained from scratch with the I3A dataset and was used to perform image classification on the ICPR 2012 and the I3A datasets. In their experiments, the authors observed the followings. First, the DA plays a crucial role in training the high-performing CNN for CL-HEP2IC. They obtained a significant boost of classification performance when DA with rotations was applied during training. Second, the background of the HEp-2 cells is useful for the classification. They trained a CNN with the mask of HEp-2 cell images and found that its performance is even lower than that of the CNN trained on HEp-2 images without taking advantage of the masks. Third, the combined prediction of CNNs trained at different epochs may give better classification performance than that of a single CNN.

In the subsequent years, \citep{jia2016deep} proposed a customized CNN model for CL-HEP2IC. The proposed model shares similarity with the VGG-M network \citep{simonyan2014very}, but uses more convolution operations. Furthermore, it uses dropout \citep{srivastava2014dropout} to overcome the effect of over-fitting. The proposed model manages to achieve competitive performance as the previous model used in \citep{gao2014hep} with a similar cost of DA (i.e., the training images are rotated at a smaller angle, e.g., 18$^\circ$ for the I3A dataset.) It is expected that the proposed network could be further improved with the use of even more convolution and batch normalizations (BN) operations \citep{ioffe2015batch}.

A recent work by \citep{ebrahim2018performance} also proposed customized models. In particular, they proposed two CNN models, one with a few layers and the other with more layers. Their models share similarity with \citep{jia2016deep}'s model, but comparatively use a smaller number of layers. In their experimental investigation on network training, they have performed thorough experiments on various pre-processing techniques. One of the interesting findings in their study is the online DA (also called as `on the fly DA'). Contrary to the traditional DA which performs DA operations on the training set before training, the online DA performs DA operations during training. However, the reported experimental results show that online DA significantly decreases the training performance. The possible reason could be the stochastic use of online DA at each training iteration of the training. This could affect the convergence of the optimization of the CNN, unless the training process is carefully managed. Meanwhile, online DA could be beneficial if used wisely. For example, the training process can begin with a certain online DA and remain unchanged until the network is converged. After that, a new online DA could be implemented to train the network further to enhance the learned features. This process can continue until the validation loss becomes stable. 

\begin{figure}[h]
    \centering
    \includegraphics[width=0.65\textwidth]{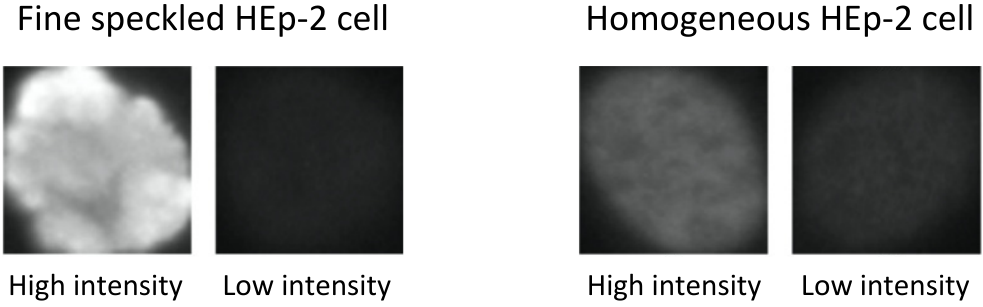}
    \caption{High and low intensity images of Fine-speckled and Homogeneous HEp-2 image classes in SNPHEp-2 dataset \citep{wiliem2013classification}. The high intensity images contain stronger cellular shape and higher illumination level, and the low intensity images contain weaker cellular shape and lower illumination level. Best viewed in soft copy.}
    \label{fig: homogeneous}
\end{figure}

Another recent work proposed by \citep{vununu2019dynamic} uses a four-stream CNN to learn local intensity and geometric information to deal with the heterogeneity problem occurring in HEp-2 cells. The intensity variations between the HEp-2 images sometimes characterize severe intra-class variations. Figure \ref{fig: homogeneous} shows images of two common HEp-2 classes, namely, fine speckled and homogeneous. For each class, high and low intensity images are shown. The high intensity images have stronger cellular shape and higher illumination level. On the other hand, the low intensity images have weaker cellular shape and lower illumination level. Note that shape and illumination are essential features for accurate classification of cells. Deterioration of shape and illumination in the low intensity images may significantly affect the discriminative capability of these features and cause serious confusion during the classification process. The proposed CNN by \citep{vununu2019dynamic} extracts features from the following discrete wavelet transform (DWT) \citep{shensa1992discrete} images, namely, horizontal detail coefficients, vertical detail coefficients, diagonal detail coefficients, and approximation coefficients with its four streams to deal with the above problem. The detail and approximate coefficients of DWT are useful in minimizing the divergences between the images with high and low intensities. The detail coefficients can be regarded as the gradients in different directions. They provide a comprehensive understanding of the cellular shapes regardless of the intensity levels present in the original image. On the other hand, the approximation coefficient provides only a certain homogenization based on the intensity level present in the original image. Learning of the above coefficients in parallel enables the proposed network to extract useful features for the classification of HEp-2 images with different cellular shapes and intensity levels. 

Similar to DWT, other image transformation methods such as fast Fourier transform, Haar wavelet transform, and directional gradient images could be used for analyzing the statistical distribution of image pixels. Also, binarized descriptors such as local binary pattern \citep{ahonen2006face} and local phase quantization \citep{ojansivu2008blur} could be used for this purpose. A further improvement of \citep{vununu2019dynamic}'s method could be to use a shared single CNN stream to process the multiple coefficient images by treating them as a set of feature channels. Using this architecture could effectively reduce the total number of network parameters and this helps improve the generalization capability of the network.

\textbf{Summary of discussion.} A large number of CL-HEP2IC methods use DNN as an end-to-end classifier. The initial methods are based on the generic DNNs that are designed for traditional image classification such as ImageNet. Motivated by the excellent results, some researchers have put efforts on the development of customized DNN architectures for CL-HEP2IC. Customized DNNs can extract more discriminative features than the generic DNNs. In terms of the DNN architectures, CNN is the most popular choice in the literature. However, some recent methods have also used CAE for feature extraction.

Existing DNNs are prone to image rotation, cells structural deterioration, and illumination variation in the image. Although theoretically, pre-alignment, uniform-scaling, and image enhancement of cell samples seem to be the solution to the aforementioned problems, they may not be effective in practice. As an alternative, most of the existing methods use DA to generate additional training data to make the CNN learn features robust against these variations. However, with the special training mechanism such as that in \citep{lei2018deeply}, it is also possible to avoid excessive usage of DA such as rotating images at a very small angle. The DA is further discussed in Section \ref{sec: pre-processing}.

From the perspective of DNN architecture, shallow CNN models are computationally efficient, but not so robust to the aforementioned variations. On the other hand, deeper CNNs such as ResNet have been successfully trained from scratch by using limited samples in cell image benchmark datasets, with the  minimum use of DA, e.g., \citep{lei2018deeply} and \citep{vununu2019deep,vununu2019dynamic}. Recent CL-HEP2IC methods are mostly based on the deeper CNN and CAE models. It is expected to see this ongoing trend in the near future.

\begin{sidewaystable}
\scalebox{0.726}{%
\begin{tabular}{lllllllll}\hline
& \multirow{2}{*}{\textbf{Reference}} & \multicolumn{2}{l}{\textbf{Pre-processing}}& \multirow{2}{*}{\textbf{Classifier}} &
\multirow{2}{*}{\textbf{{Dataset}}} &
\multirow{2}{*}{\textbf{\makecell[l]{{Dataset Split/Used}\\{Evaluation Protocol}}}} &
\multirow{2}{*}{\textbf{Results (\%)}} & \multirow{2}{*}{\textbf{Remarks}} \\ \cline{3-4}
&&\textbf{EN}&\textbf{DA}&&&\\ \hline
\multirow{3}{*}[-2.7ex]{\rotatebox[origin=c]{0}{\textbf{SCP-SL-HEP2IC}}} & \cite{li2016deep} & $\Cross$     & R, M   & Modified LeNet-5 & {I3A Task-1} & {LOSO} & 83.55 & \makecell[l]{Use simple network, cross-dataset based training,\\[-0ex] prone to illumination variation.} \\ \cline{2-9}
& \cite{li2016hep} & $\Cross$ & R & \makecell[l]{Modified LeNet-5\\ + SVM} & {I3A Task-2} & {LOSO} & 85.62 & \makecell[l]{Simple network, cell population histogram,\\[-0ex]  prone to illumination and intra-class variations.} \\ \cline{2-9} 
 & \cite{cascio2019deep} & CS & R & AlexNet+SVM & {I3A Task-1} & {LOSO} & 93.75 & \makecell[l]{Two-stage framework, class-specific feature fusion,\\[-0ex] multi-stage training, use segmentation mask.}   \\ \hline
\multirow{4}{*}[-4.0ex]{\rotatebox[origin=c]{0}{\textbf{MCP-SL-HEP2IC}}} & \cite{li2016hep2} & CS & R, C & FCN & {I3A Task-2} & {LOSO} & 90.89 & \makecell[l]{Single-shot prediction, multi-task network,\\[-0ex] low-resolution segmentation map.}   \\ \cline{2-9}
& \cite{li2017hep} & CS & R, M, C & Extended FCRN & {I3A Task-2} & {LOSO} & 94.67 & \makecell[l]{Efficient network architecture, generalized features,\\[-0ex] low resolution segmentation map.}   \\ \cline{2-9}
& \cite{oraibi2018learning} & $\Cross$ & R & \makecell[l]{VGG-19 + LBP\\ + JML} & {I3A Task-2} & {LOSO} & 92.11 & \makecell[l]{Hybrid features, efficient framework,\\[-0ex]  handcrafted features, prone to illumination variation.}   \\ \cline{2-9}
& \cite{xie2019deeply} & $\Cross$  & R, M, C & DSFCN & {I3A Task-2} & {LOSO} & 95.40 & \makecell[l]{Rich feature map, parameterized fusion,\\[-0ex]  prone to illumination variation and blur.} \\ \hline
\end{tabular}}
\caption{Summary of deep HEp-2 specimen classification methods on the I3A Task-1 and Task-2 dataset. The widely accepted evaluation protocol for the I3A dataset is MCA with the LOSO (leave-one-specimen-out) protocol (refer to Section 4 for details). EN = Image enhancement method; CS = Contrast stretching; DA = Data augmentation; R = Rotation; F = Flipping; C = Cropping; M = Mirroring; SCP-SL-HEP2IC = Single-cell processing based SL-HEP2IC methods; MCP-SL-HEP2IC = Multi-cell processing based based SL-HEP2IC methods.}
\label{tab: sl-hep2ic}
\end{sidewaystable}

\subsection{Specimen-level HEp-2 image classification (SL-HEP2IC) methods}
\label{sec: SL-HEP2IC}
Unlike the CL-HEP2IC methods, the SL-HEP2IC methods perform classification of the whole HEp-2 specimen images. Based on how the specimen image is processed to obtain the classification results, existing SL-HEP2IC methods can be further decomposed into two major types, namely, single-cell processing based SL-HEP2IC methods (SCP-SL-HEP2IC) and multi-cell processing based methods (MCP-SL-HEP2IC). Table \ref{tab: sl-hep2ic} provides a summary of both types of methods. It organizes these methods based on their year of publication in an ascending order. Furthermore, for each method, the table lists the following key information: pre-processing techniques, basic DNN architectures, performance on the benchmark datasets, and remarks. Sections \ref{sec: SCP-SL-HEP2IC} and \ref{sec: MCP-SL-HEP2IC} provide a thorough discussion on them.

\subsubsection{Single-cell processing based specimen-level HEp-2 image classification methods (SCP-SL-HEP2IC)}
\label{sec: SCP-SL-HEP2IC}
Given a specimen image, SCP-SL-HEP2IC methods decompose it into individual cell images by cell-level ground truth labels, i.e., bounding box annotations and segmentation masks. Next, each individual cell image is classified by a DNN. The classification results of each of the cell images are then accumulated to obtain the specimen-level result (e.g., via the majority voting strategy). SCP-SL-HEP2IC methods can also be regarded as the extension of CL-HEP2IC methods (discussed in Section \ref{sec: CL-HEP2IC}) to the SL-HEP2IC. Based on the use of DNN, SCP-SL-HEP2IC methods can be further divided into two types, namely, DNN feature extraction based methods and true or pure DNN based methods (i.e., methods that use the DNN as an end-to-end classifier).

\begin{figure*}[t]
    \centering
    \includegraphics[width=0.9\textwidth]{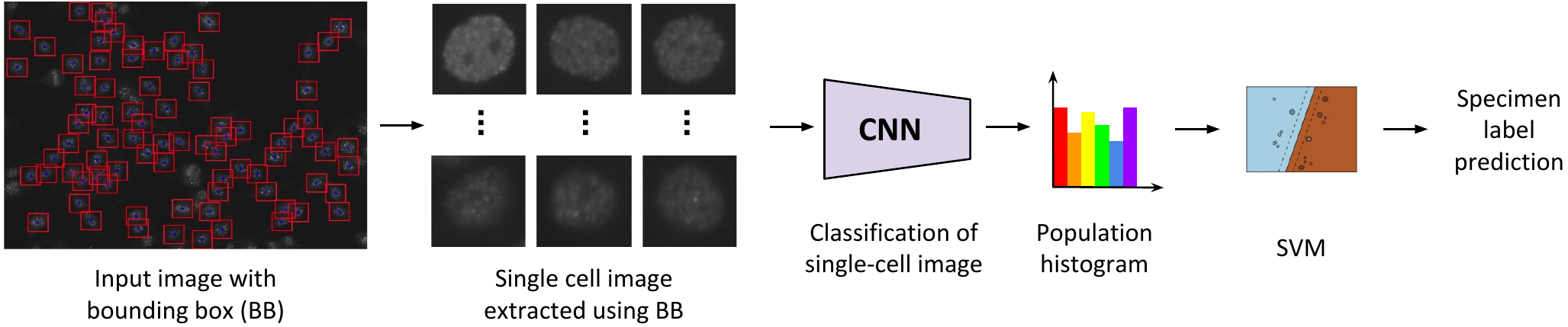}
    \caption{Overview of the HEp-2 population histogram proposed by \protect\citep{li2016hep}. Given a specimen image, the individual HEp-2 cell images are extracted using the bounding box annotations. A CNN is then used to classify the individual cells. Based on the population of the predicted cell labels in the specimen, the cell population histogram (PH) is then constructed. Finally, an SVM is used for the classification of PH. Better viewed in color.}
    \label{fig: pattern histogram}
\end{figure*}

\textbf{Feature extraction based methods.} Similar to the methods discussed in Section \ref{sec: CL-HEP2IC-DNN-FE}, the DNN feature extraction based methods use a two-stage pipeline for the classification of individual cell images obtained from the input specimen image. In the first stage, the DNN features are extracted from each individual cell image. Following that, in the second stage, the extracted features are classified using a traditional classifier, e.g., SVM. \citep{li2016hep} proposed one of the very few methods to consider the extraction of DNN features for individual cell classification. They have used a modified LeNet-5 CNN model for feature extraction. Specifically, a few $1\times 1$ convolutional blocks are added to the original LeNet-5 architecture to increase the depth of the feature channels. The modified LeNet-5 CNN model gives better performance than the original LeNet-5 model. Once the classification of individual cell images is done, a majority voting (MV) strategy is applied to the predicted cell labels to obtain the specimen label. The MV strategy selects the most dominant cell label as the specimen label, which works well on all the I3A Task-2 dataset, except the \textit{Mitotic Spindle} class. A majority of the \textit{Mitotic Spindle} specimens are misclassified as the \textit{Golgi} and \textit{Homogeneous} specimens. To solve this issue, \citep{li2016hep} proposed to represent each specimen by a population histogram (PH). A PH is a simple histogram that describes the frequency of existing individual cells in a specimen image. Figure \ref{fig: pattern histogram} shows the pipeline for PH construction. From the point of view on features, it can be regarded as a high-level feature representation. The experimental results of \citep{li2016hep} demonstrate that the PH based strategy classifies the \textit{Mitotic Spindle} images more accurately than the MV strategy. 

Since the PH is based on the individual cell image prediction by the DNN, adoption of a recent DNN model such as ResNet could help improve its discriminative capacity. Furthermore, in the proposed LeNet-5 based CNN model, \citep{li2016hep} assumed that it can well handle the illumination variations, and did not use any image enhancement techniques. However, in other HEP2IC methods such as \citep{gao2017hep}, image enhancement is used on LeNet-5 to obtain better invariance across the illumination variations. Hence, it is believed that the proposed LeNet-5 based CNN model by \citep{li2016hep} could be further improved with the use of image enhancement.

In a recent work by \citep{cascio2019deep}, the authors used a simpler approach to classify individual cell images extracted from a specimen sample. The basic working principle of this method is discussed as a part of CL-HEP2IC, and interested readers are referred to Section \ref{sec: CL-HEP2IC-DNN-FE}. As a side note, unlike \citep{li2016hep}'s method, the reported performance of \citep{cascio2019deep}'s method is based on the I3A Task-1 dataset which is relatively small. It is worth mentioning that the I3A Task-1 dataset is originally proposed for CL-HEP2IC and contains only individual cell images. However, it also provides the specimen ID for each cell image. The specimen IDs can be used to identify the membership of each cell image with respect to different specimen samples. The SCP-SL-HEP2IC methods leverage this information to perform SL-HEP2IC and evaluate the performance with the widely used leave-one-specimen-out (LOSO) scheme. At each step of this scheme, all the cell images belonging to one specimen are reserved for testing and the cell images belonging to the remaining specimens are used for training. The main drawback of this method is that it uses pre-trained features from an early CNN model \citep{krizhevsky2012imagenet}. The use of fine-tuned features from a more recent CNN model such as ResNet may further improve the performance of this method.

\textbf{True DNN based methods.} True or pure DNN based SCP-SL-HEP2IC means that those methods employ the DNN as a classifier. They are very similar to the methods discussed in Section \ref{sec: CL-HEP2IC-DNN-CLFR}. An early approach is proposed by \citep{li2016deep}. They have employed a modified LeNet-5 CNN model to classify the individual cell images extracted from specimen images. Their modified LeNet-5 CNN model is similar to the model proposed by \citep{li2016hep}. Even though the proposed CNN is simple, it achieves a good performance on the I3A Task-1 dataset. The key to their good performance is the use of cross-dataset samples for training. Readers are referred to Section \ref{sec: cross-dataset} for the details of cross-dataset training.

\textbf{Segmentation of individual cell images.} Segmentation of individual cell images from a specimen image plays a vital role in achieving good SCP-SL-HEP2IC performance. A straightforward approach to this could be the use of bounding box annotations which segment cells with the cell-surrounding regions. Another alternative is the use of segmentation mask which exactly segments a cell out, without including any cell-surrounding region. \citep{cascio2019deep} performed an investigation on both approaches and found that the latter approach performs slightly better than the former. However, this investigation was based on pre-trained ImageNet features from an early CNN model, i.e., AlexNet. It is believed that the use of a CNN model that has been fine-tuned with HEp-2 cell images may minimize this resulting gap. Prior to \citep{cascio2019deep}, \citep{li2016hep,li2016deep} used segmentation masks for individual cell segmentation but they did not conduct further investigation on other cell segmentation methods such as bounding box. In addition, in the work of \citep{gao2017hep}, the authors recommended the inclusion of cell-surrounding regions to train CNNs. They found that the use of a segmentation mask leads to inferior classification performance. However, \citep{gao2017hep}'s observation was for CL-HEP2IC. Moreover, in SCP-SL-HEP2IC, the classification results of individual cells are combined by a majority vote. As long as the dominating cell pattern matches with the label of the specimen, the specimen will be classified correctly.

\textbf{Summary of discussion.} Feature extraction based methods are more popular for SCP-SL-HEP2IC. The key disadvantage of the SCP-SL-HEP2IC methods is their requirement of thorough classification of single cells in the specimen image to obtain the specimen label. The SCP-SL-HEP2IC methods are more suitable for the cases where only a small number of single cells are present in a specimen image.

\subsubsection{Multi-cell processing based specimen-level HEp-2 image classification methods (MCP-SL-HEP2IC)}
\label{sec: MCP-SL-HEP2IC}
Unlike the SCP-SL-HEP2IC methods described in Section \ref{sec: SCP-SL-HEP2IC}, MCP-SL-HEP2IC based methods process an entire specimen image at a time. Hence, they are computationally more efficient. Based on their nature, they can be further decomposed into two types, namely, pixel-wise prediction based methods and image-wise prediction based methods.

\begin{figure}[!h]
    \centering
    \includegraphics[width=1\textwidth]{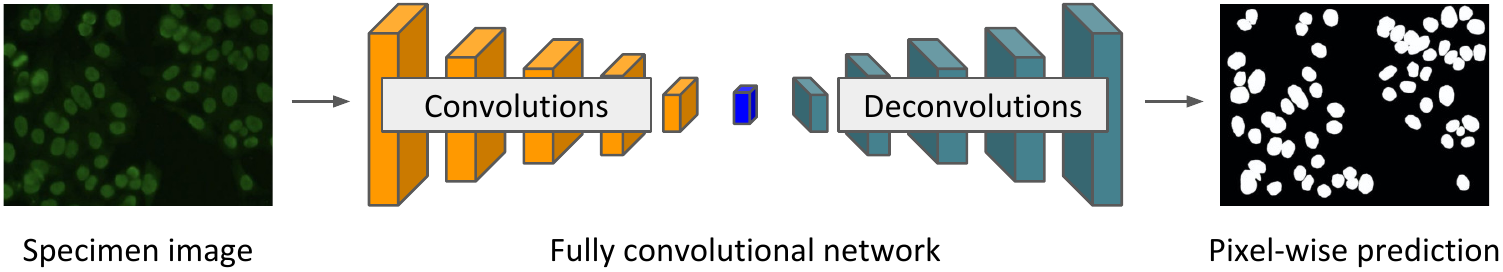}
    \caption{Fully convolutional network (FCN) \protect\citep{long2015fully} used in the method proposed by \citep{li2016hep2}. An FCN consists of convolution and deconvolution layers. The convolution layer learns the high-level feature representation, and the deconvolution layer reconstructs the fine details for semantic segmentation.}
    \label{fig: fcn}
\end{figure}

\textbf{Pixel-wise prediction based methods.} Pixel-wise prediction based methods perform dense prediction on a specimen image such that each pixel of the specimen image is given a class label. The classification of the specimen is performed via an MV strategy on the predicted pixel labels. Due to the dense predictions, pixel-wise prediction based methods are capable of solving multiple tasks such as HEp-2 image classification and segmentation. The fully convolutional network (FCN) \citep{long2015fully} is often used as a backbone in pixel-wise prediction based methods. The FCN is a variant of CNN that replaces the fully connected layers with convolution layers. Figure \ref{fig: fcn} shows the architecture of FCN used for the task of MCP-SL-HEP2IC.

\citep{li2016hep2} are the first to consider FCN for SL-HEP2IC. They simply adopted the FCN model proposed by \citep{long2015fully} for the task and managed to achieve good classification performance on the I3A Task-2 dataset. FCN shows the following key advantages: (1) it can process a specimen image in a single-shot, which saves a significant amount of computation time during both training and testing; (2) as opposed to the CNN, it can operate on arbitrary input sizes; hence, fixation of input specimen image size during training and testing is not necessary; and (3) it can make dense predictions which could be further used for cell segmentation. However, FCN suffers from low-resolution feature maps. Due to the propagation of an input image through a stack of layers composed of convolution and pooling operations, FCN cannot provide high resolution (i.e., very precise object boundaries) feature maps. Low-resolution feature maps may not well represent cells captured at a smaller scale and can cause fuzzy object boundaries.

\begin{figure}[!h]
    \centering
    \includegraphics[width=1\textwidth]{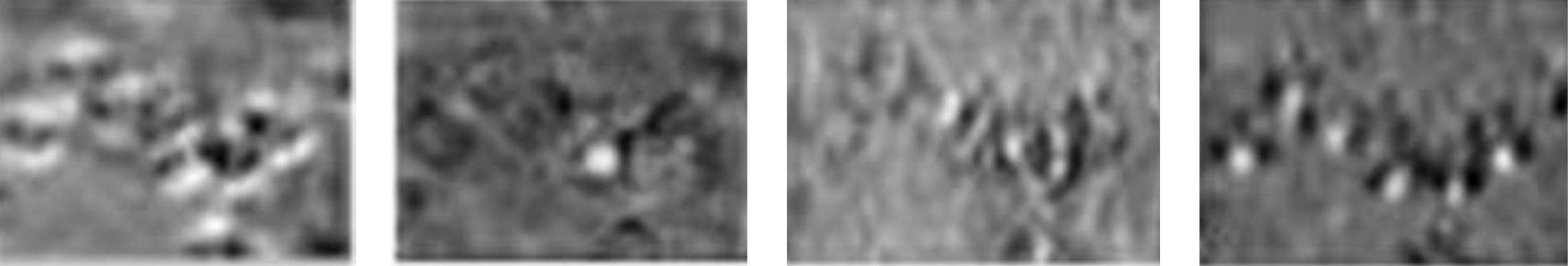}
    \caption{Feature maps produced by the fourth RiR module in the proposed extended FCRN network \protect \citep{li2017hep}. The feature maps show that the RiR module is capable of learning effective features from HEp-2 specimen images. Image courtesy of \citep{li2017hep}.} 
    \label{fig:rir-maps}
\end{figure}

Following the success of FCN, in a subsequent work by \citep{li2017hep}, the authors employed a modified fully convolutional residual network (FCRN) \citep{wu2016high} to further improve the classification performance. The authors extend the original FCRN by increasing its depth to nearly double (1.76 times) by replacing the bottleneck module with the residual in residual (RiR) module. The RiR module has more layers and identity shortcuts than the bottleneck module. Typically, the depth of FCRN is increased by adding more residual modules which are computationally expensive, i.e., more trainable parameters. Unlike this, the RiR modules allow the FCRN to increase its depth without requiring more residual modules. The proposed extended FCRN outperforms the baseline FCRN-50 and FCN, and achieves a good performance (i.e., 94.67\%) on the I3A dataset. Figure \ref{fig:rir-maps} shows some example feature maps learned by the RiR module in the proposed network. The extended FCRN is an effective and efficient approach for specimen classification. However, there are some cell classes such as \textit{homogeneous} and \textit{mitotic spindle} for which the proposed network suffers to achieve better classification accuracy. This could be due to their higher class similarities which confuse the classification.

A more recent work by \citep{xie2019deeply} proposed a simple yet effective method to generate high-resolution feature maps by the FCN for SL-HEP2IC. They proposed to combine the downsampled feature maps in the convolution layers with the upsampled feature maps in the deconvolution layers. However, unlike in the original FCN, they proposed to use the skip connections between intermediate layers to preserve the rich object boundary details. Furthermore, a parametric feature fusion strategy was also proposed to better fuse the intermediate layer features. Experimental results on the I3A Task-2 dataset demonstrate that the proposed method outperforms the above two methods by \citep{li2017hep}. However, \citep{xie2019deeply} indicated that the proposed method is not robust against illumination variations and blurry image conditions. 

\textbf{Image-wise prediction based methods.} Given a specimen image, the image-wise prediction based methods return a single label for the whole input image. A recent method proposed by \citep{oraibi2018learning} considers a simple approach to SL-HEP2IC. Unlike the pixel-wise prediction based SL-HEP2IC methods, this method resizes the specimen image and classifies it with a VGG-19 CNN model \citep{simonyan2014very}. Since the resizing operation deteriorates the local cell information, the authors proposed to use handcrafted features such as local binary pattern \citep{ojansivu2008blur} and joint motif labels \citep{prasath2016hep} to strengthen the discriminative capacity of CNN features. The experimental result shows that the proposed method surpasses the one proposed by \citep{li2016hep2}. The followings could be used to improve the performance of the proposed method: (1) more discriminative CNN features such as ResNet-based features instead of VGG-based features and (2) gradient-based local shape features such as SIFT \citep{lowe1999object} instead of motif labels. 

\textbf{Summary of discussion.} MCP-SL-HEP2IC methods are computationally more efficient than the SCP-SL-HEP2IC methods due to their multiple cell processing capacity. They are also more practical and scalable. While the pixel-wise prediction based methods are the more intensively researched MCP-SL-HEP2IC methods, image-wise prediction based methods are also finding their way to SL-HEP2IC. However, due to the multi-tasking capability of pixel-wise prediction based methods, i.e., segmentation and classification, it is expected that more SL-HEP2IC research will be conducted along this direction.

\subsection{Positive and Negative HEp-2 Specimen Image Classification}

In the clinical setting, HEP2IC is conducted in two main phases. In the first phase, HEp-2 specimen images are classified into positive or negative categories. The images in the negative category are considered to be `negative' for the ANA test. On the other hand, images in the positive category are considered to be `positive' for the ANA test and are taken to the second phase for further analysis. A large amount of research has been conducted on ANA positive category image classification such as the CL-HEP2IC and SL-HEP2IC methods. Unfortunately, research on HEp-2 positive and negative image classification is still limited.

The early methods developed for this classification task such as \citep{wiliem2011automatic}, \citep{flessland2002performance} and \citep{hiemann2010method} were sensitive to the image acquisition procedures and struggled to deal with multiple staining patterns \citep{hobson2016computer}. Comparatively, more recent methods are less sensitive to image acquisition and have a better ability in handling multiple staining patterns. Rather than relying on complex image acquisition procedures, they use advanced visual features to deal with various illumination conditions and staining patterns in the HEp-2 specimen images. For example, \citep{di2016analyte} used local contrast features at multiple scales to characterize the image intensity. \citep{benammar2016computer} used a mixture of intensity, geometry and shape features. \citep{zhou2017positive} used global color and local gradient features with a two-stage classification framework. \citep{cascio2019automatic} used an optimally selected mixture of intensity and texture features with linear discriminant analysis \citep{ye2005two}. \citep{merone2019computer} used texture-like features produced by an invariant scattering convolutional network \citep{bruna2013invariant}. Recently, deep learning based features have also been used for this classification task. For example, \citep{cascio2019intensity} used image features from pre-trained CNN models. Among those methods, the classification results obtained by \citep{zhou2017positive} and \citep{cascio2019intensity} are most promising. In particular, \citep{zhou2017positive} achieved 98.7\% accuracy on the SZU HEp-2 cell dataset \citep{zhou2017positive} and \citep{cascio2019intensity} achieved 92.8\% accuracy on the AIDA (autoimmunity: diagnosis assisted by computer) dataset \citep{benammar2016computer}. It is believed that the adoption of features from a fine-tuned CNN model would further improve the performance of \citep{cascio2019intensity}'s method. Also, it is expected that more research involving deep learning will be conducted in the future to address this classification task.

\subsection{Common Discussion}

This section will present discussions on the CL-HEP2IC and SL-HEP2IC methods in Sections \ref{sec: CL-HEP2IC} and \ref{sec: SL-HEP2IC}, respectively.

\subsubsection{Pre-processing techniques used in CL-HEP2IC and SL-HEP2IC methods}
\label{sec: pre-processing}
Pre-processing techniques have been widely used in the existing deep learning based HEP2IC methods. A list of pre-processing techniques that have been employed is available in Tables \ref{tab:cell-level methods} and \ref{tab: sl-hep2ic}. Based on the type of operations, these techniques can be divided into two groups, namely, image enhancement techniques and data augmentation techniques. Both are discussed as follows.

\textbf{Image enhancement:} Image enhancement (IE) is a common pre-processing technique used in many computer vision tasks. In the existing deep learning based HEP2IC methods, IE has been used as an image normalization technique. Among many IE techniques, contrast stretching (CS) which scales the range of intensity values of an input image to a desired range of values is widely used. There are some papers such as \citep{bayramoglu2015human}, \citep{rodrigues2017exploitingnpreprocessing}, and \citep{rodrigues2017hep} that consider other types of IE techniques such as histogram equalization. It is worth mentioning that the use of IE is popular among the early HEP2IC methods. Recent HEP2IC methods use more deep models and do not require IE to handle image intensity variations.

\textbf{Data augmentation:} DA is widely used in the existing deep learning based HEP2IC methods. Rotations (R), mirroring (M), flipping (F), and cropping (C) remain the most popular DA operations. These operations produce transformed images while keeping the semantic information in the original images. To obtain images with different compositions from the original image, GAN can be used to produce synthesized cell images \citep{majtner2019effectiveness}. The use of GAN for DA in HEP2IC is still in its primary stage (to the best of our knowledge, only one work in this area has used GAN based DA \citep{majtner2019effectiveness}). DA can be applied to the HEp-2 datasets in two ways, i.e., (1) conventional way, and (2) class-aware way. In the conventional way, the DA operation is performed on every image classes of the HEp-2 datasets without looking at its distribution. The conventional way is not very effective for datasets that have small tails in its distribution. Comparatively, the class-aware way is more effective for datasets with imbalanced class distributions such as the I3A dataset. Specifically, it applies additional DA operations to the classes at the tail of the distribution to make the class sizes more balanced. The class-aware DA is widely used in the existing HEP2IC methods.

\subsubsection{Evaluation strategies used in CL-HEP2IC and SL-HEP2IC methods}
The common evaluation strategies used for the CL-HEP2IC and SL-HEP2IC methods are average classification accuracy (ACA) and mean classification accuracy (MCA), respectively. Both of these strategies are discussed in details in Section \ref{sec: datasets}. In the case of CL-HEP2IC, all the datasets are splited into training and test sets (optionally, validation set) in different ratios (more discussions are provided on this in the following section). Differently, the SL-HEP2IC commonly uses leave-one-specimen-out (LOSO) evaluation protocol for training and testing (i.e., possibly due to a limited number of samples available in the existing datasets). In addition, a few SL-HEP2IC methods also report their performance using the $k$-fold cross validation.

\section{HEp-2 Public Datasets}
\label{sec: datasets}
\begin{figure*}[t]
    \centering
    \includegraphics[width=1\textwidth]{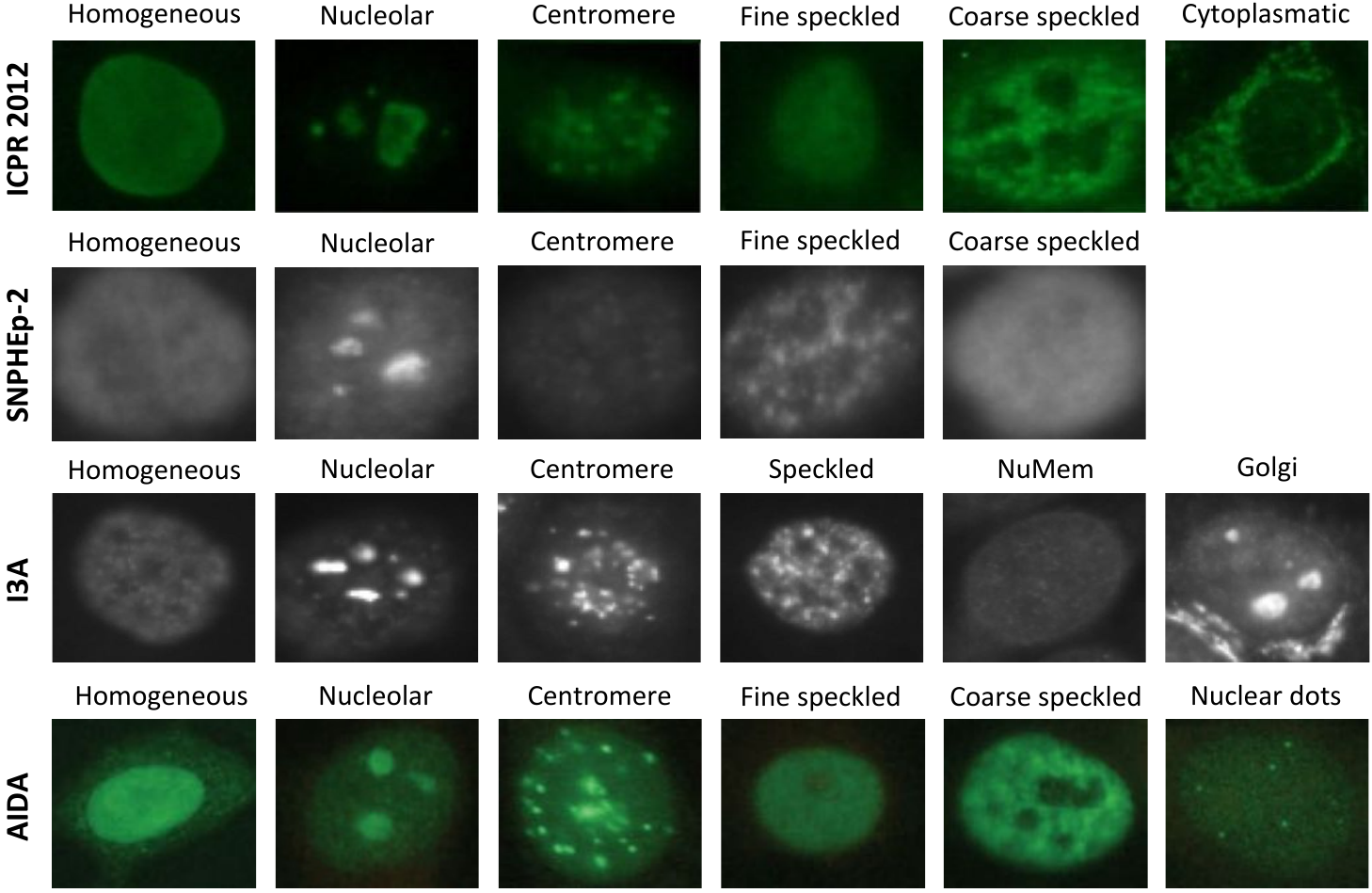}
    \caption{Sample categories available in existing HEp-2 public cell image datasets.}
    \label{fig: datasets}
\end{figure*}

Similar to other cell image datasets, the compilation of a dataset that contains HEp-2 cell images is costly, requiring special image acquisition equipment and expert judgments on data annotations. Due to this, at this moment, there are only a few datasets available publicly for the development of HEP2IC approaches. Figure \ref{fig: datasets} shows a sample of various HEp-2 cell image classes in existing datasets.  The performance of the existing HEP2IC methods on various HEp-2 public datasets is given in Table \ref{tab:cell-level methods} and Table \ref{tab: sl-hep2ic}. This section provides a review on the existing public HEp-2 datasets and their evaluation strategies widely used by the community.

\subsection{ICPR 2012 dataset}
The ICPR 2012 dataset\footnote{Download link for the ICPR 2012 dataset: \url{https://mivia.unisa.it/contest-hep-2/}} is also known as the MIVIA HEp-2 images dataset, named after the MIVIA laboratory of the University of Salerno as the first public dataset released during `Contest on HEp-2 Cells Classification' at ICPR 2012 \citep{foggia2013benchmarking}. It has a total of 1,455 individual cell images distilled from twenty-eight 1,388$\times$1,038 pixels color specimen samples and partitioned into 721 and 734 images for training and testing, respectively. Immunology experts annotate each specimen at cell level. Based on the annotations, each cell image is classified into one of the following six categories, namely,  \textit{Homogeneous, Coarse speckled, Fine speckled, Nucleolar, Centromere,} and \textit{Cytoplasmic}. Figure \ref{fig: datasets} shows example images from these categories. For performance evaluation of this dataset, ACA proposed by \citep{foggia2013benchmarking} is widely used in current HEP2IC methods. The ACA is calculated as follows:

\begin{equation*}
\text{ACA}=\frac{\text{\#Number of correctly predicted samples}}{\text{\#Number of total samples}}
\end{equation*}

\subsection{SNPHEp-2 dataset}
The SNPHEp-2 dataset\footnote{Download link for the SNPHEp-2 dataset: \url{http://staff.itee.uq.edu.au/lovell/snphep2}}, in short, the SNPHEp-2 dataset \citep{wiliem2013classification}, is another public dataset used in the literature. In terms of the image classes, it has a higher similarity with the ICPR 2012 dataset and the AIDA dataset than the I3A dataset. The only difference between the SNPHEp-2 and the I3A datasets is that SNPHEp-2 dataset does not have the \textit{Nuclear Membrane} and \textit{Golgi} classes. There are in total 1,884 individual monochrome cell images available in this dataset, from which 905 images are selected for training and 979 images are selected for testing purpose. A five-fold cross-validation strategy is applied to evaluate the performance of this dataset. Figure \ref{fig: datasets} shows example images of this dataset. In comparison with the ICPR 2012 dataset, the SNPHEp-2 dataset is less used in the existing HEP2IC methods.

\subsection{I3A dataset}

The I3A dataset\footnote{Download link for the I3A dataset: \texttt{https://hep2.unisa.it/dbtools.html}} is one of the commonly used dataset in the literature. It was introduced in the `competition on cells classification by fluorescent image analysis' held at ICIP (International Conference on Image Processing) 2013 \citep{hobson2013competition} and subsequently reused in I3A classification competitions hosted by ICPR (International Conference on Pattern Recognition) 2014 \citep{hobson2015benchmarking} and ICPR 2016 \citep{lovell2016international}. There exist two versions of this dataset, namely, Task-1 and Task-2 datasets. Task-1 is primarily designed for CL-HEP2IC and the Task-2 is designed for SL-HEP2IC. Both Task-1 and Task-2 datasets are divided into the training and the test sets. During the release of Task-1 and Task-2 datasets for the competitions, only the training sets are made available to the public by organizers. The test set is kept as private and used by the organisers for evaluating the performance of the participating methods in the competitions.

The Task-1 training set has a total of 13,596 monochrome single cell images extracted from 83 specimens using the bounding box annotations by experts. The images are divided into six classes, namely, \textit{Homogeneous, Speckled, Nucleolar, Centromere, Nuclear membrane (NuMem)}, and \textit{Golgi}, as shown in Figure \ref{fig: datasets}. Unlike the ICPR 2012 and the SNPHEp-2 datasets, the \textit{Coarse speckled} and \textit{Fine speckled} classes are represented using a single class called \textit{Speckled} class. The \textit{NuMem} and \textit{Golgi} are new and not available in the above datasets. Since there is no test set publicly available for this dataset, existing deep learning based CL-HEP2IC methods use one of the following strategies, (1) split the training set into 64\% (8,701 images), 16\% (2,175 images), and 20\% (2,720 images) of the samples and use them for training, validation, and test purposes \citep{gao2017hep}, (2) split the training set into 80\% (10,876 images) and 20\% (2,720 images) of the samples and use them for training and test purposes, e.g., \citep{jia2016deep} and \citep{li2018hep},  and (3) k-fold cross validation, e.g., \citep{han2016hep} and \citep{rodrigues2017hep}.

\begin{figure}[h]
    \centering
    \includegraphics[width=1\textwidth]{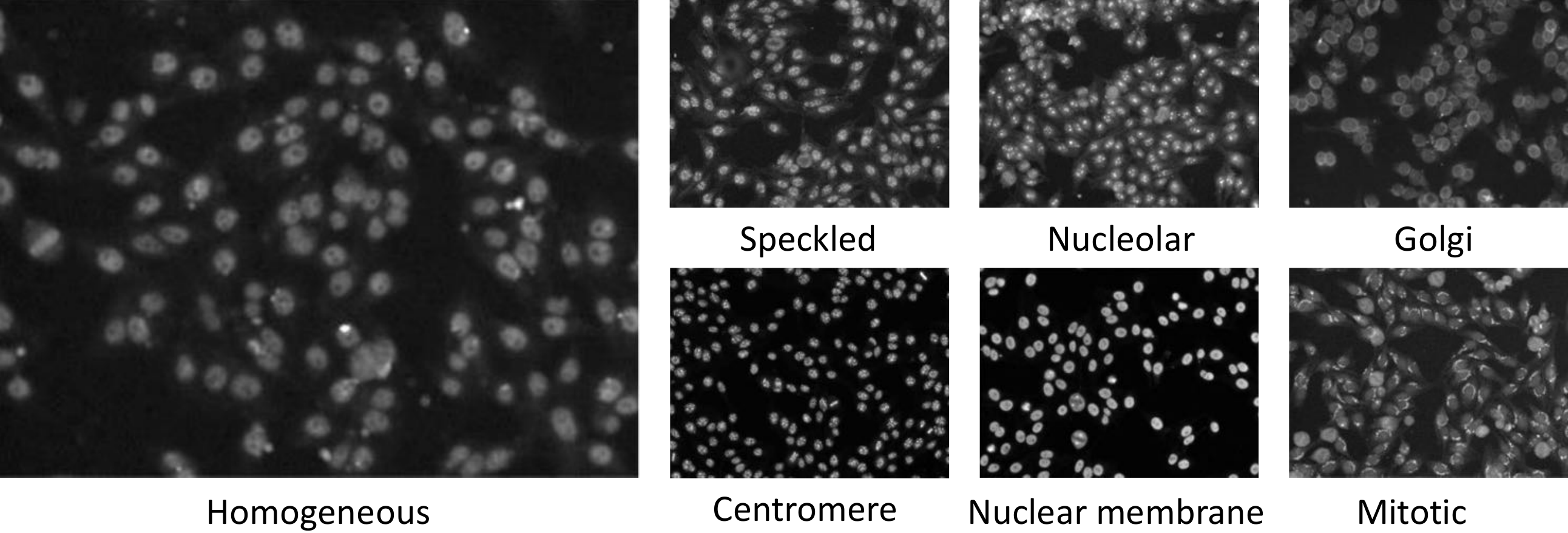}
    \caption{Sample specimen images of various classes in the I3A Task-2 dataset.}
    \label{fig:icpr14-task2}
\end{figure}

The Task-2 dataset has a total of 1,008 images, and they are taken from four different locations of 252 specimen samples, i.e., 252$\times$4 = 1,008. The images in this dataset are distributed across seven classes, namely, \textit{Speckled} (208 images), \textit{Nucleolar} (200 images), \textit{Homogeneous} (212 images), \textit{Mitotic spindle} (60 images), \textit{Golgi} (40 images), \textit{Nuclear membrane} (64 images), and \textit{Centromere} (204 images). Figure \ref{fig:icpr14-task2} shows the sample specimen images from each class. In the I3A and the ICPR 2016 competitions, the participants trained their models with the provided training set and submitted them to the organizers for testing. Similar to the Task-1 dataset, the test set of the Task-2 dataset is not released to the public. 
As previously mentioned, existing SL-HEP2IC methods reviewed in this paper widely apply the leave-one-specimen-out (LOSO) protocol to the training set for benchmarking. 

MCA is generally used for performance measure of both the Task-1 and Task-2 datasets. ACA is used by only a selective methods. MCA is calculated by taking the mean of all class accuracies:

\begin{equation}
\textnormal{MCA}= \frac{1}{C}\sum_{i=1}^{C}\textnormal{CA}_i
\end{equation}
where $\textnormal{CA}_i$ is the classification accuracy of class $i$ and $C$ is the total number of image classes. The $\textnormal{CA}$ is defined as follows:

\begin{equation*}
\text{CA}=\frac{\text{\#Number of correctly predicted samples}}{\text{\#Number of total samples in a class}}
\end{equation*}

\subsection{AIDA dataset}
The autoimmunity: diagnosis assisted by computer (AIDA) dataset\footnote{Download link for the AIDA dataset: \texttt{http://www.aidaproject.net/index.php/it/downloads}} \citep{benammar2016computer} is a large-scale HEp-2 image dataset proposed as part of the AIDA project. The AIDA dataset is divided into two parts, namely, private and public. The AIDA private dataset has 20,000 samples, but it is only available to AIDA project partners. The AIDA public dataset has 2,080 specimen images and each specimen has (single or multiple) HEp-2 cells from 22 types of antinuclear antibody patterns including those in the previous datasets. The image size is not fixed and only specimen-level annotation by human experts is provided. One major difference of AIDA from the previous datasets is that it has negative intensity samples while others have only positive and weak positive samples. The positive, weak positive, and negative intensity images have high, medium, and low intensity levels, respectively. Specifically, AIDA has 582 negative intensity samples and 1,498 positive intensity samples. Until now, this dataset has only been used for the HEp-2 specimen fluorescence intensity classification \citep{cascio2019intensity}. As a result, no evaluation protocol (e.g., the training, validation, and test image sets) exists for the HEp-2 pattern classification on this dataset. However, following the data partitioning scheme used in the I3A dataset, the AIDA dataset may be divided into training, validation, and test splits for the HEP2IC task. Figure \ref{fig: datasets} shows some examples of the HEp-2 cell patterns in the AIDA public dataset. 

\subsection{Cross-HEp-2 dataset recognition}
\label{sec: cross-dataset}
Designing a high-performing classifier is challenging. In the conventional classification tasks, test samples are always expected to have the same statistical characteristics as training samples. However, in the case of cross-dataset classification, the classifier is trained with one dataset and tested on another dataset that could have a different data distribution. The classifier is expected to handle the change in data distribution and give a good classification performance. Existing HEp-2 datasets are not produced under the same laboratory settings, so the sample varies in terms of contrast, illumination variation, scale, and rotation. To perform cross-dataset classification, a classification system should handle these issues well. One way to deal with this situation is to combine all the samples from the existing datasets to create a super training set, and use it to train the classifier \citep{bayramoglu2015human,li2016deep}. Meanwhile, in the existing literature, for cross-dataset recognition, most of the works train a classifier with a larger sized source dataset and then fine-tune it with a target dataset \citep{li2018hep,li2017hep,li2016hep,li2016deep,liu2017hep,shen2018deep}. Usually, the I3A dataset is used as a source dataset due to its relatively larger size than the other datasets. However, since I3A dataset is still not large enough and existing deep networks require a substantial amount of data for successful training, a few works proposed effective training strategies such as cross-modal transfer learning \citep{lei2018deeply} to successfully train deep networks with small datasets. It is expected that advanced techniques such as domain adaptation \citep{wang2018deep} and transfer learning \citep{zhang2017transfer} would be exploited in further to deal with the issues of cross-HEp-2 dataset recognition. 

\section{Discussions and Future Trends}
\subsection{Overview}
This paper has reviewed the literature of deep learning based HEP2IC until October 2019. There are a total of 31 papers currently available that use DL for HEP2IC. Although the use of DL began in 2013, most of the papers were published in the last four years (2016-2019). At the early stage, the use of DL was only limited to CL-HEP2IC. However, in the recent years, DL has been successfully applied to SL-HEP2IC. The networks used in the existing CL-HEP2IC and SL-HEP2IC methods are covered. Only a few methods (see Section \ref{sec: CL-HEP2IC-DNN-FE}) currently use pre-trained deep networks as feature extractor, which can be regarded as the extension of handcrafted methods. The majority of the methods (see Tables \ref{tab:cell-level methods} and \ref{tab: sl-hep2ic}), especially those published in the last five years, use end-to-end trainable deep architectures. It can be confidently said that the use of such networks has become an established practice. 

\subsection{Key features of most successful deep HEP2IC methods}
After conducting the review of the existing literature of CL-HEP2IC and SL-HEP2IC, we have noticed that designing a perfect network architecture may not be the research focus in achieving higher HEp-2 cell image classification performance. It has been seen that even with the established general-purpose deep networks, some authors have managed to achieve the state-of-the-art results \citep{lei2017cross}. Scrutinizing the methods that obtain good classification performance, we can see that they mostly go beyond the culture of just deepening the network by adding more layers. In particular, they look into the techniques of deep learning such as data augmentation and data enhancement \citep{rodrigues2017exploitingnpreprocessing}.  At the same time, data augmentation and enhancement are not the only keys to achieve good performance. Some authors have suggested to use special mechanisms such as multi-scale feature extraction \citep{li2018hep,li2017deep}, knowledge sharing between layers \citep{lei2018deeply}, and training with transformed images \citep{vununu2019deep,vununu2019dynamic}.

\subsection{Unique challenges of HEp-2 image classification}
HEP2IC using DL faces some special challenges. In the existing literature, the limitation of sample size in the current HEp-2 datasets is often mentioned as a great challenge. However, the recent studies \citep{lei2018deeply,li2018hep} show that it may be only partially valid. With the efficient design of deep architecture or a little tweak in generic DNN models, one could effectively train a high-performing HEP2IC system with smaller datasets. Instead, the greater challenge with the current datasets is dealing with various imaging conditions. Although the current datasets are collected in laboratories under controlled conditions, their images present significant challenges for deep networks. The acquisition of HEp-2 datasets is carried out in several stages, and usually, in batches. Due to the manual staining process, the samples among different batches vary in terms of lighting conditions and staining pattern strengths (the strengths of HEp-2 staining pattern depend on the serum sample which varies from person to person). Also, the HEp-2 datasets are often collected in a collaborative manner. Many research groups from different geographic locations actively participate in the image acquisition, where each group collects distinctive blood serum samples for the HEp-2 staining. The HEp-2 staining patterns collected from these samples may significantly increase the variation of the the dataset. Most of the existing methods deal with the various image acquisition conditions by applying image enhancement techniques such as contrast stretching. However, some recent methods have considered other techniques such as gradient image \citep{vununu2019deep} and DWT transformation \citep{vununu2019dynamic}. 

Another challenging issue in the existing HEp-2 cell image datasets lies at that they may not contain a similar number of samples across all classes. Some classes have a smaller number of samples, i.e., tail classes, than most of the other classes. Training DNNs with such datasets could focus too much on learning robust features for the dominant classes. The minor classes receive less attention during training and this may adversely affect their classification. One of the effective solutions proposed by the existing methods to this problem is class-balanced data augmentation \citep{li2017deep}. It is expected that more efficient solutions would be developed with the research progress on HEP2IC.

Lastly, the labeling of HEp-2 images is another challenging task. There are two types of labeling currently used in the existing datasets, namely, cell-level labeling and specimen-level labeling. Usually, several immunology experts take part in the HEp-2 image labeling process to achieve observer-independent annotations since it is very common to label the same HEp-2 image differently by different experts \citep{foggia2013benchmarking,hobson2015benchmarking}. This is a time-consuming process and usually takes 2-3 years for a dataset \citep{foggia2013benchmarking,lovell2016international,wiliem2013classification}. The ICPR 2012, SNPHEp-2, and I3A datasets have cell-level annotations, whereas the AIDA dataset only has specimen-level annotations. Although the labeling challenge is primarily concerned with HEp-2 dataset collection and compilation, it has a significant impact on the development of DL system for HEP2IC.

\subsection{Future trends of deep learning in HEp-2 cell image classification}
Despite the great research progress in recent years, there are still limitations
in the existing methods to overcome in the future following the ongoing progress of DL research. Currently, there are only four datasets available publicly for HEP2IC. Although the existing HEP2IC methods use data augmentation to increase the samples in the datasets, we must agree that these data augmentations are largely basic image transformations and they are not efficient in exploring the true distribution of HEp-2 cell images. Synthesized image generation (e.g., with the GAN technique) could be a better solution to this issue. However, synthesized image generation for HEP2IC is still in its beginning stage. To the best of our knowledge, at present, only one work has explored the use of GAN based DA for HEP2IC \citep{majtner2019effectiveness}. It is expected that more research along this line will be explored. Meanwhile, it is hoped that in the future more datasets could be released for public use. For example, following the success of recent HEP2IC competitions \citep{foggia2013benchmarking,hobson2015benchmarking}, new competitions using large-scale HEp-2 image datasets could be further organized to encourage researchers to continue to focus on the HEP2IC issues.

From the network design perspective, the methods published between 2017 and 2019 are slightly different from the earlier ones. They are more focused on reducing the number of network parameters \citep{li2017deep}, designing the networks to be trainable with smaller datasets \citep{li2018hep}, and avoiding the excessive use of data pre-processing \citep{lei2018deeply}. It is expected that the trend of designing computationally efficient HEP2IC methods will continue to grow in the future. 

Labeling images is one of the most challenging tasks during the compilation \citep{foggia2013benchmarking, hobson2015benchmarking} of HEp-2 cell image datasets. Weakly-supervised, self-supervised, and unsupervised learning are the three promising areas of deep learning which could help overcoming this challenge. While weakly-supervised methods require some initial label information, self-supervised and unsupervised methods could be conducted without label information. One of the unsupervised learning methods, CAE, has been successfully employed in HEP2IC for unsupervised feature learning, and it shows promising results \citep{liu2017hep,vununu2019deep}. In the coming years, it is expected that more research will be conducted in these areas to address the labeling limitation.

At this moment, deep learning methods are largely treated as `black boxes' in the literature. As indicated by \citep{litjens2017survey}, having just a good classification system is not sufficient in medicine. There are substantial risks of legal issues involved in the whole process. Existing HEP2IC methods are limited in this aspect. It is expected that in the future more methods will be developed to understand, interpret and explain the response of various parts of deep networks and the final classification decision with respect to HEp-2 cell images. 

Lastly, consistency of results is another issue and most of the existing HEP2IC methods do not sufficiently consider the issue of result consistency in deep networks. The optimization of deep networks is based on stochastic procedures which generally give more focus on the performance and could undermine the consistency of the produced results \citep{piantadosi2018reproducibility}. It is hoped that in the future, statistical significance analysis shall be extensively conducted on the deep network results to verify the consistency. Such an analysis would help practitioners to better understand the results produced by the deep networks.

From the above discussions, we can conclude that deep learning has dramatically impacted the progress of the HEP2IC and more exciting results are waiting ahead.

\section*{Acknowledgments}
This work was supported by the CSIRO Data61 Scholarship; the University of Wollongong IPTA Scholarship; the Australian Research Council (grant number DE160100241); and the Multi-modal Australian ScienceS Imaging and Visualisation Environment (MASSIVE) (www.massive.org.au). The authors are also grateful to the anonymous reviewers for their insightful comments that have improved the manuscript.

\bibliographystyle{apalike}
\bibliography{references}

\end{document}